\documentclass[sigconf]{acmart}
\usepackage{amsthm}
\newtheorem{proposition}{Proposition}[section]
\newtheorem{remark}{Remark}[section]

\usepackage{algorithm}
\usepackage{algorithmic}
\usepackage{graphicx}
\usepackage{svg}
\usepackage{multirow}
\usepackage{enumitem} 
\usepackage{tabularx}
\usepackage{float}
\usepackage{longtable}
\usepackage{subcaption} 
\usepackage{enumitem}
\usepackage{rotating}
\usepackage{balance}

\AtBeginDocument{%
  }

\copyrightyear{2025}
\acmYear{2025}
\setcopyright{acmlicensed}\acmConference[KDD '25]{Proceedings of the 31st ACM SIGKDD Conference on Knowledge Discovery and Data Mining V.2}{August 3--7, 2025}{Toronto, ON, Canada}
\acmBooktitle{Proceedings of the 31st ACM SIGKDD Conference on Knowledge Discovery and Data Mining V.2 (KDD '25), August 3--7, 2025, Toronto, ON, Canada}
\acmDOI{10.1145/3711896.3736989}
\acmISBN{979-8-4007-1454-2/2025/08}

\begin{document}

\title{Graph Evidential Learning for Anomaly Detection}

% Yunhai Wang, Yueguo Chen, Bing Bai, Fei Wang
\author{Chunyu Wei}
\affiliation{%
  \institution{Renmin University of China}
  \city{Beijing}
  \country{China}
}
\email{weicy15@icloud.com}

\author{Wenji Hu}
\affiliation{%
  \institution{Renmin University of China}
  \city{Beijing}
  \country{China}
}
\email{2024000991@ruc.edu.cn}

\author{Xingjia Hao}
\affiliation{%
  \institution{Guangxi University}
  \city{Nanning}
  \country{China}
}
\email{haoxingjia@st.gxu.edu.cn}

\author{Yunhai Wang}
\authornote{Corresponding author.}
\affiliation{%
 \institution{Renmin University of China}
 \city{Beijing}
 \country{China}}
\email{cloudseawang@gmail.com}

\author{Yueguo Chen}
\affiliation{%
  \institution{Renmin University of China}
  \city{Beijing}
  \country{China}}
\email{chenyueguo@ruc.edu.cn}

\author{Bing Bai}
\affiliation{%
  \institution{Microsoft MAI}
  \city{Beijing}
  \country{China}}

\email{bingbai@microsoft.com}

\author{Fei Wang}
\affiliation{%
  \institution{Cornell University}
  \city{New York}
  \country{United States}}
\email{few2001@med.cornell.edu}

\renewcommand{\shortauthors}{Chunyu Wei et al.}
\begin{abstract}
Graph anomaly detection faces significant challenges due to the scarcity of reliable anomaly-labeled datasets, driving the development of unsupervised methods. 
Graph autoencoders (GAEs) have emerged as a dominant approach by reconstructing graph structures and node features while deriving anomaly scores from reconstruction errors. 
However, relying solely on reconstruction error for anomaly detection has limitations, as it increases the sensitivity to noise and overfitting.
To address these issues, we propose Graph Evidential Learning (GEL), a probabilistic framework that redefines the reconstruction process through evidential learning. 
By modeling node features and graph topology using evidential distributions, GEL quantifies two types of uncertainty: graph uncertainty and reconstruction uncertainty, incorporating them into the anomaly scoring mechanism. Extensive experiments demonstrate that GEL achieves state-of-the-art performance while maintaining high robustness against noise and structural perturbations.

\end{abstract}




\maketitle

\section{Introduction}
\label{sec:introduction}
Anomaly detection focuses on identifying objects that significantly deviate from the majority within a dataset~\cite{DBLP:journals/csur/ChandolaBK09}. With the rapid proliferation of relational data driven by the Internet, graph-structured data has become a natural representation for modeling complex, interconnected systems. This has led to growing interest in Graph Anomaly Detection (GAD), which aims to detect anomalous nodes within large-scale graphs. GAD has diverse applications in fields such as fraud detection~\cite{DBLP:journals/ipm/ZengLJHL25}, network intrusion prevention~\cite{DBLP:journals/compsec/DengH25}, and the identification of abnormal behaviors in social networks~\cite{DBLP:journals/corr/SavageZYCW16}, biological systems~\cite{DBLP:journals/sensors/GrekovKVT23}, and financial transactions~\cite{DBLP:conf/kdd/ChenT22}. 

Despite its practical importance, labeled data in graph settings is often scarce compared to the vast scale of interaction data, making supervised approaches infeasible in most scenarios. Manual labeling of anomalies is not only expensive and time-consuming but also impractical due to the high diversity and rarity of anomalous behaviors. These have driven the adoption of unsupervised methods for GAD, enabling adaptive and scalable solutions that can generalize across various anomaly types and datasets.

Existing unsupervised methods for GAD predominantly rely on \textbf{reconstruction-based approaches}, where autoencoders are commonly employed to learn low-dimensional representations of graph data. The central hypothesis is that autoencoders capture the core structure of normal nodes, while anomalies, due to their sparsity and distinctiveness, are poorly reconstructed, resulting in higher reconstruction errors. \textbf{GAEs}, which incorporate Graph Neural Networks (GNNs) to encode both topological structures and node attributes, have shown promise in detecting anomalies within large, complex graphs~\cite{DBLP:conf/nips/LiuDZDHZDCPSSLC22}. 

\begin{figure}[!t]
  \centering
  \includegraphics[width=0.99\linewidth]{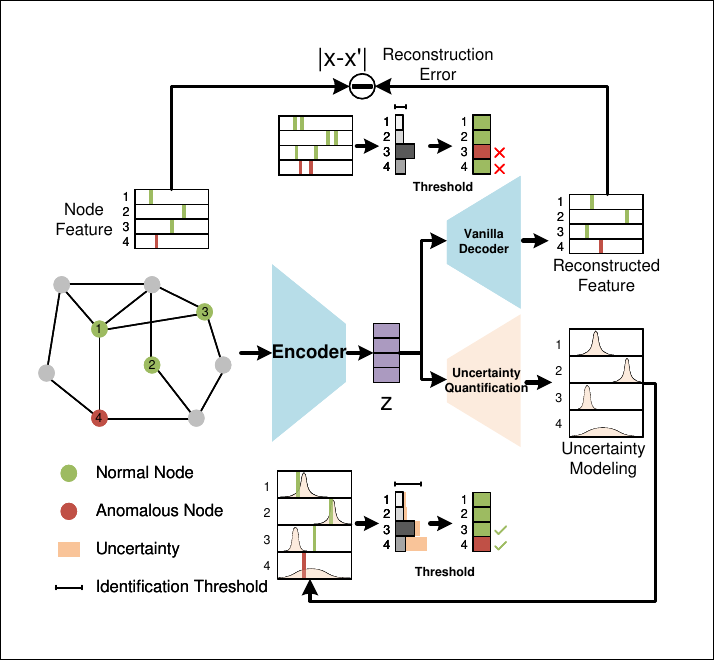}
  \vspace{-2ex}
  \caption{Introducing Uncertainty for anomaly detection. Relying on reconstruction error, nodes 3 and 4 in the graph are misclassified without introducing uncertainty.}
  \label{fig:motivation}
  \vspace{-2ex}
\end{figure}

However, existing anomaly detection methods based on reconstruction error exhibit several limitations, as illustrated in Figure~\ref{fig:motivation}.
On one hand, most reconstruction-based methods assume that normal nodes can be accurately reconstructed, while anomalous nodes cannot. However, in real-world scenarios, these methods are prone to overfitting, allowing anomalous nodes to be reconstructed with high accuracy, thereby evading detection. 
On the other hand, these methods typically rely on predefined reconstruction error threshold to distinguish between normal and anomalous nodes. In practical applications, this approach is highly sensitive to inherent noise in graph data and reconstruction models, such as inconsistencies in node features or errors in structural encoding. As a result, high-noise normal nodes may fail to be accurately reconstructed, leading to their misclassification as anomalies.

To overcome these challenges, we propose shifting the paradigm of graph anomaly detection from relying on reconstruction error to explicitly modeling uncertainty. This approach aims to address the shortcomings of sensitivity to noise and overfitting, enabling the model to express high uncertainty when encountering anomalies.
However, integrating uncertainty into graph reconstruction introduces two fundamental challenges:
\begin{itemize}[leftmargin=*]
    \item \textbf{Uncertainty Diversity.} There exist diverse sources of uncertainty in reconstruction: (1) Graph uncertainty, which arises when anomalies disguise themselves by interacting frequently with normal nodes or mimicking their features, creating inconsistencies in local topological information; (2) Reconstruction uncertainty, which occurs when the reconstruction model, trained predominantly on normal data, encounters anomalies that lie outside the training distribution. Effective anomaly detection requires a unified approach to capture both sources of uncertainty.
    \item \textbf{Modality Heterogeneity. } Graph data inherently combines \textbf{continuous node features} and \textbf{discrete topological structures}, each following distinct distributional characteristics. Node features are often modeled as continuous distributions in high-dimensional spaces, while topological structures are discrete and binary, representing edge presence or absence. Estimating uncertainty for these heterogeneous modalities requires differentiated modeling strategies while simultaneously capturing their interdependencies within the unified graph structure.
\end{itemize}
% \vspace{-0.8ex}

To address the aforementioned challenges simultaneously, we propose Graph Evidential Learning (GEL), a novel framework that integrates uncertainty modeling for graph anomaly detection. 

To address \textbf{Uncertainty Diversity}, we adopt a higher-order evidential distribution to parameterize the reconstruction process. Instead of separately modeling different sources of uncertainty (e.g., using Monte Carlo Dropout or training multiple models to estimate prediction variance~\cite{DBLP:conf/nips/AbeBPZC22}), we directly learn a unified evidential distribution for graph reconstruction. This approach eliminates the need for task-specific uncertainty estimation techniques, enabling a more efficient and generalizable anomaly detection framework. The higher-order evidential distribution generates a set of lower-order reconstruction likelihood functions, thus encapsulating both graph and reconstruction uncertainty simultaneously. By leveraging Bayesian inference, we parameterize this evidential distribution without requiring extensive sampling, as typically required by Bayesian Neural Networks.

% To address \textbf{Modality Heterogeneity}, we introduce a joint evidential distribution that simultaneously models uncertainties in continuous node features and discrete topological structures. Specifically, we represent the reconstruction likelihood of continuous node features using a Normal distribution, and model the reconstruction likelihood of discrete topological structures (e.g., edge presence) using a Categorical distribution. These likelihood functions are jointly governed by a higher-order evidential distribution, allowing us to capture the interdependencies between features and structure in a principled manner. We use neural network to parameterize this evidential distribution, enabling efficient sampling of reconstruction likelihoods for both modalities.
To address \textbf{Modality Heterogeneity}, we introduce two evidential distributions to model the uncertainties in continuous node features and discrete topological structures, respectively. Specifically, we employ a Normal Inverse-Gamma (NIG) distribution to represent the reconstruction distribution of continuous node features and a Beta distribution to model the reconstruction distribution of discrete topological structures (i.e., the presence or absence of edges between nodes). These two higher-order evidential distributions jointly govern the reconstruction of the graph, enabling us to capture the interdependencies between features and structures in a principled manner. Neural networks are used to parameterize these evidential distributions, allowing efficient sampling of the reconstruction likelihoods for both modalities.

Our key contributions can be summarized as follows:
% \vspace{-1.5ex}
\begin{itemize}[leftmargin=*]
    \item We propose a novel paradigm for GAD by shifting from reconstruction error to uncertainty modeling, enhancing the robustness of anomaly detection in noisy graph data. To comprehensively capture the various sources of uncertainty, we model uncertainty using a higher-order evidential distribution.
    \item We handle the diverse modality of node features and topological structures through a joint evidential distribution.
    \item Extensive experiments on five datasets demonstrate that GEL achieves state-of-the-art performance in GAD. Ablation studies show that GEL is notably more robust in unsupervised settings.
\end{itemize}

\section{Related Works}
\label{sec:related_work}
\subsection{Graph Anomaly Detection}
Anomaly detection has been applied in fields like financial fraud detection~\cite{hilal2022financial} and network security~\cite{togay2021firewall}. Traditional methods, including statistical techniques and distance-based approaches like k-nearest neighbors (k-NN)\cite{li2007network}, isolation forests\cite{ding2013anomaly}, and PCA\cite{huang2006network}, detect anomalies based on statistical deviations but struggle with high-dimensional or non-Euclidean data~\cite{pang2021deep}.
Deep learning approaches, such as autoencoders~\cite{memarzadeh2020unsupervised}, VAEs\cite{li2020anomaly}, and RNNs\cite{nanduri2016anomaly}, capture patterns in high-dimensional data but overlook the relational structure in graphs, limiting their anomaly detection capability.

GNNs~\cite{DBLP:conf/iclr/KipfW17} excel at utilizing graph structure for anomaly detection~\cite{kim2022graph}. For example, Mul-GAD~\cite{liu2022mul} integrates node attributes and structure, while \citet{kumagai2021semi} address class imbalance using semi-supervised GCNs. AddGraph~\cite{zheng2019addgraph} leverages temporal GCNs for dynamic GAD. However, these methods often require substantial labeled data, which is scarce~\cite{ma2021comprehensive}.
Unsupervised approaches, such as structure-based methods~\cite{DBLP:conf/kdd/XuYFS07}, distance-based methods like Node2Vec~\cite{DBLP:conf/kdd/GroverL16}, and subgraph-based techniques~\cite{DBLP:journals/tifs/YuanSY23}, detect anomalies by analyzing graph topology and local subgraphs. Spectral and density-based methods, like LOF~\cite{DBLP:conf/sigmod/BreunigKNS00}, detect anomalies based on graph irregularities~\cite{DBLP:journals/tkde/HuGLWDM20}.
Deep learning-based methods, including GAEs~\cite{DBLP:journals/corr/KipfW16a}, reconstruct graph structures to detect anomalies based on reconstruction errors~\cite{DBLP:conf/icassp/FanZL20}. Variants like VGAEs~\cite{DBLP:journals/corr/KipfW16a} and GraphGANs~\cite{DBLP:conf/aaai/ZhengYW0L19} enhance robustness. Methods such as DOMINANT~\cite{DBLP:conf/sdm/DingLBL19} and GAD-NR~\cite{DBLP:conf/wsdm/RoySLYES024} extend GAEs by reconstructing node neighborhoods.

However, traditional GAEs face two main issues: reliance on reconstruction error, making them sensitive to noise, and inability to handle overfitting samples, leading to misclassification. We address these challenges by introducing uncertainty as an anomaly criterion to improve robustness and reliability.

\subsection{Uncertainty Quantification}
Uncertainty Quantification (UQ) has become essential for improving the reliability and interpretability of neural network predictions~\cite{charpentier2020posterior}, especially in safety-critical areas like autonomous driving~\cite{michelmore2018evaluating}, healthcare~\cite{nemani2023uncertainty}, and finance~\cite{blasco2024survey}.
Bayesian Neural Networks (BNNs) were an early approach to UQ by assigning prior distributions to model parameters and enabling posterior inference~\cite{blundell2015weight, DBLP:conf/mm/00080ZW0J022, DBLP:journals/eswa/YongB22}. However, their practical use is limited by high-dimensional inference intractability, computational costs of methods like variational inference, and challenges in choosing priors with limited domain knowledge~\cite{fortuin2021bnnpriors}.

Ensemble methods, including MC Dropout~\cite{gal2015dropout} and Deep Ensembles~\cite{lakshminarayanan2017simple}, offer an alternative by leveraging diverse predictions from multiple models. While effective, they incur high computational overhead and face challenges in maintaining ensemble diversity, particularly in resource-limited settings.

Evidential Learning provides a more efficient UQ alternative by directly modeling uncertainty via higher-order distributions over predictions~\cite{sensoy2018evidential}. In regression, it uses a Normal-Inverse-Gamma distribution to model Gaussian parameters~\cite{amini2020deep}, and in classification, it employs the Dirichlet distribution for class probabilities~\cite{sensoy2018evidential}. Unlike BNNs and ensembles, evidential learning enables efficient uncertainty estimation in a single forward pass, making it suitable for real-time applications. It has been applied successfully in image classification~\cite{sensoy2018evidential}, regression~\cite{amini2020deep}, multi-view learning~\cite{liu2024merit}, and OOD detection~\cite{cui2022out}.

Building on this, we introduce evidential learning to unsupervised graph representation learning by modeling graph topology and node features as a joint higher-order evidential distribution, enhancing uncertainty assessment in graph reconstruction.

\begin{figure*}[!t]
  \centering
  \includegraphics[width=0.99\linewidth]{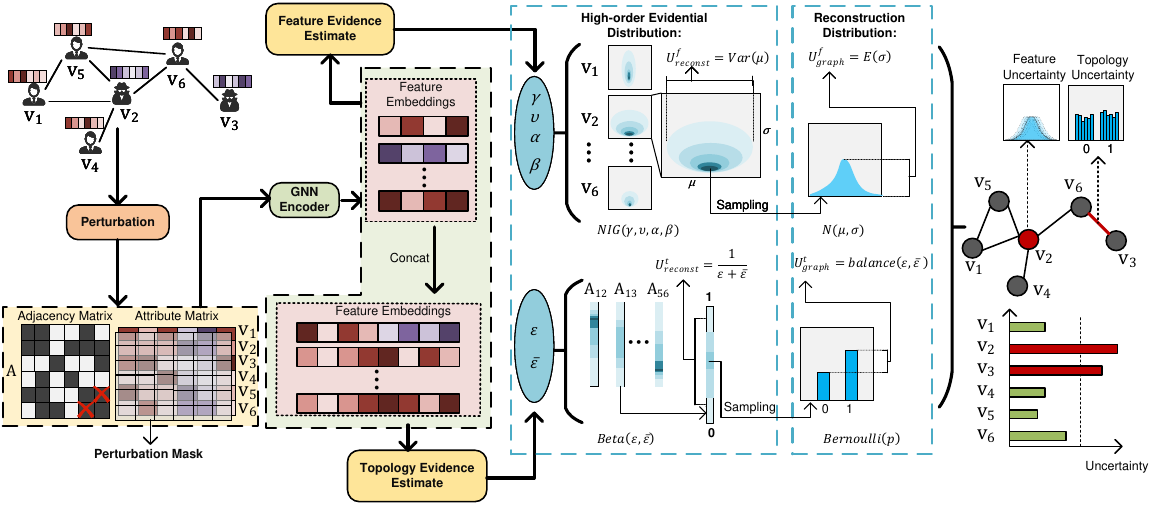}
  \vspace{-2ex}
  \caption{The overview of the GEL framework. GEL models high-order evidential distribution to calculate feature uncertainty and topology uncertainty during the graph reconstruction process, leveraging these uncertainties for anomaly detection.}
  \vspace{-1ex}
  \label{fig:main}
\end{figure*}
\section{Preliminaries}
\label{sec:preliminaries}

\subsection{Problem Definition}
\paragraph{Attributed Graph.}
An \textit{attributed graph} consists of nodes, their attributes, and the relations (edges) among them~\cite{DBLP:conf/wsdm/RoySLYES024}. Formally, $\mathcal{G} = \{\mathcal{V}, \mathcal{E}, \mathbf{X}\}$, where $\mathcal{V} = \{v_1, v_2, \ldots, v_N\}$ is the set of $N$ nodes, $\mathcal{E} = \{e_{ij} \mid v_i \text{ and } v_j \text{ are connected}\}$ is the edge set, and $\mathbf{X} \in \mathbb{R}^{N \times d}$ is the feature matrix, where each row $\mathbf{x}_i \in \mathbb{R}^d$ represents the $d$-dimensional attributes of node $v_i$.
The edge set $\mathcal{E}$ can be represented by an \textit{adjacency matrix} $\mathbf{A} \in \mathbb{N}^{N \times N}$, where $\mathbf{A}_{ij} = 
1$, if $e_{ij} \in \mathcal{E}$, and 0 otherwise.
The \textit{degree matrix} $\mathbf{D} \in \mathbb{N}^{N \times N}$ is a diagonal matrix where $\mathbf{D}_{ii} = \sum_{j=1}^{N} \mathbf{A}_{ij}$, and $\mathbf{D}_{ij} = 0, \, \forall i \neq j$.
Here, $\mathbf{D}_{ii}$ denotes the degree of node $v_i$, with off-diagonal entries being zero.

\paragraph{Unsupervised Graph Anomaly Detection.}
Given an \textit{attributed graph} $\mathcal{G}$, the node set $\mathcal{V}$ is divided into two disjoint subsets: the anomalous node set $\mathcal{V}_a$ and the normal node set $\mathcal{V}_n$, such that $\mathcal{V}_a \cap \mathcal{V}_n = \emptyset$ and $\mathcal{V}_a \cup \mathcal{V}_n = \mathcal{V}$. The goal of unsupervised graph anomaly detection is to assign anomaly labels for all nodes by estimating the probability $p(v_i \in \mathcal{V}_a \mid \mathcal{G}, v_i \in \mathcal{V})$, which represents the likelihood that node $v_i$ is anomalous, given the graph $\mathcal{G} = \{\mathcal{V}, \mathcal{E}, \mathbf{X}\}$.

\subsection{Reconstruction with GAE}
\label{sec:GAE}
An autoencoder (AE) compresses high-dimensional data into a latent representation via an encoder and reconstructs it through a decoder. AEs capture normal data properties, with large reconstruction errors indicating anomalies. This extends to GAEs, where the encoder uses a GNN to incorporate node features and topology. Given a graph $\mathcal{G} = \{\mathcal{V}, \mathcal{E}, \mathbf{X}\}$, the encoder is:
{\small
\begin{equation}
\mathbf{Z} = \text{Enc}(\mathcal{G}) = \text{GNN}(\mathbf{X}, \mathbf{A}),
\end{equation}
}where $\mathbf{Z} \in \mathbb{R}^{N \times d'}$ is the latent representation, with each $\mathbf{Z}_i \in \mathbb{R}^{d'}$ corresponding to node $v_i \in \mathcal{V}$, and $d' \ll d$.
To capture higher-order neighbor information, GNNs stack multiple graph convolution layers. At layer $l$, node embeddings $\mathbf{H}^{(l)}$ are updated by aggregating neighbor information:
{\small
\begin{equation}
\mathbf{H}^{(l+1)} = \sigma\left( \mathbf{\tilde{A}} \mathbf{H}^{(l)} \mathbf{W}^{(l)} \right),
\end{equation}
}where $\mathbf{\tilde{A}} = \mathbf{D}^{-\frac{1}{2}} (\mathbf{A} + \mathbf{I}) \mathbf{D}^{-\frac{1}{2}}$ is the normalized adjacency matrix with self-loops, and $\mathbf{W}^{(l)}$ is the weight matrix. The output $\mathbf{H}^{(L)}$ is the compressed representation $\mathbf{Z}$.

\textbf{Feature Reconstruction. }
The decoder reconstructs node features by taking $\mathbf{Z}$ as input and outputting $\hat{\mathbf{X}} \in \mathbb{R}^{N \times d}$. The decoder $\text{Dec}_f$ is a multi-layer perceptron (MLP):
{\small
\begin{equation}
\hat{\mathbf{X}} = \text{Dec}_f(\mathbf{Z}) = \text{MLP}(\mathbf{Z}).
\end{equation}
}
% The reconstruction error is the Mean Squared Error (MSE):
% \begin{equation}
% \mathcal{L}_f = \frac{1}{N \times d} \sum_{i=1}^N \sum_{h=1}^d \left( X_{ih} - \hat{X}_{ih} \right)^2.
% \end{equation}

\textbf{Topology Reconstruction. }
To reconstruct the graph topology, the decoder predicts the adjacency matrix $\mathbf{A}$ by estimating edge presence. For nodes $i$ and $j$, the decoder $\text{Dec}_t$ computes the probability of an edge as:
{\small
\begin{equation}
\mathbf{\hat{A}} = \text{Dec}_t(\mathbf{Z}) = \sigma\left( \mathbf{Z}^\top \mathbf{Z} \right),
\end{equation}
}where $\sigma(x) = \frac{1}{1 + e^{-x}}$ is the sigmoid activation function.

We adopt Mean Squared Error (MSE) for optimizing both feature and topology reconstruction.

\section{Methodology}
\label{sec:methodology}
% As illustrated in figure~\ref{fig:main}, GEL consists of three core components. 
% In Section~\ref{sec:GAE}, we introduced the decoders $\text{Dec}_f$ and $\text{Dec}_t$ designed to reconstruct node features and graph topology from the latent embedding $\mathbf{Z}$. However, these decoders, being neural networks, do not provide a measure of their confidence in the reconstruction process. This lack of uncertainty quantification often leads to inaccurate outputs. The problem is exacerbated when anomalies resemble normal patterns, as the decoders might generate high-confidence reconstructions for these anomalous nodes or edges, thus diminishing the reliability of the detection system. To mitigate this issue, we propose a novel framework inspired by the evidential paradigm~\cite{sensoy2018evidential}, which decouples the graph reconstruction process into two distinct stages: (1) generating evidence for each graph component, and (2) deriving reconstruction results based on this evidence.
Figure~\ref{fig:main} illustrates the basic framework of GEL. In Section~\ref{sec:GAE}, we introduced the decoders $\text{Dec}_f$ and $\text{Dec}_t$ designed to reconstruct node features and graph topology from the latent embedding $\mathbf{Z}$. However, these neural network-based decoders lack a measure of confidence in the reconstruction process, which can lead to unreliable outputs. 
% This issue is exacerbated when anomalies mimic normal patterns, as the decoders might produce high-confidence reconstructions for anomalous nodes or edges, reducing the detection system's reliability.
To address this limitation, we propose an evidential framework, which decouples the graph reconstruction process into two stages: (1) generating evidence for graph components, and (2) deriving reconstruction results based on evidence.

\subsection{Learning the Evidence and Uncertainty}
\subsubsection{Feature Evidence}
In traditional graph autoencoders, decoders directly output a point estimate for each node feature, but such estimates fail to capture the inherent uncertainty of predictions. We propose modeling node feature estimates as continuous distributions to reflect this uncertainty. Specifically, we assume a Gaussian prior $\mathcal{N}_0(\boldsymbol{\mu}, \boldsymbol{\sigma}^2)$ for the node feature $\mathbf{X}$.

To model the uncertainty of the Gaussian parameters $\boldsymbol{\mu}$ and $\boldsymbol{\sigma}^2$, we adopt a second-order probabilistic framework. We place a Gaussian prior on the mean $\boldsymbol{\mu}$, and an Inverse-Gamma prior on the variance $\boldsymbol{\sigma}$, as follows:
\begin{equation*}
\boldsymbol{\mu} \sim \mathcal{N}(\boldsymbol{\gamma}, \boldsymbol{\sigma}^2 \boldsymbol{\nu}^{-1}), \quad \boldsymbol{\sigma}^2 \sim \boldsymbol{\Gamma}^{-1}(\boldsymbol{\alpha}, \boldsymbol{\beta}),
\end{equation*}
where $\mathcal{N}(\cdot)$ denotes a Gaussian distribution, and $\Gamma^{-1}(\cdot)$ refers to an Inverse-Gamma distribution. 
The hyperparameters $\boldsymbol{\gamma}$, $\boldsymbol{\nu}$, $\boldsymbol{\alpha}$, and $\boldsymbol{\beta}$ are all of the same size as the feature matrix $\mathbf{X} \in \mathbb{R}^{N \times D}$, and encode the evidence supporting the Gaussian distribution $\mathcal{N}_0$. 
Specifically, we have $\boldsymbol{\gamma} \in \mathbb{R}$, $\boldsymbol{\nu} > 0$, $\boldsymbol{\alpha} > 1$, and $\boldsymbol{\beta} > 0$. For clarity, we denote the hyperparameters corresponding to a specific node's feature vector $\mathbf{X}_{ih}$ in $\mathbf{X}$ by using their non-bold versions: $\mu$, $\sigma$, $\alpha$, $\beta$, $\gamma$, and $\nu$.

We aim to estimate the posterior distribution $q(\mu, \sigma^2) = p(\mu, \sigma^2 | \mathbf{X}_{ih})$. To approximate the true posterior~\cite{parisi1988statistical}, we assume that the distribution can be factorized as $q(\mu, \sigma^2) = q(\mu)q(\sigma^2)$, which allows us to efficiently model it using the Normal-Inverse-Gamma (NIG) distribution, a conjugate prior for the Gaussian. The joint density function of the NIG distribution is given by:
{\small
\begin{equation}
\label{equ:NIG}
p(\mu, \sigma^2 \mid \gamma, \nu, \alpha, \beta) = \frac{\beta^{\alpha} \sqrt{\nu}}{\Gamma(\alpha) \sqrt{2 \pi \sigma^2}} \left(\frac{1}{\sigma^2}\right)^{\alpha+1} \exp\left\{-\frac{2\beta + \nu(\gamma - \mu)^2}{2\sigma^2}\right\},
\end{equation}
}where $\Gamma(\cdot)$ represents the Gamma function.

The parameters of this conjugate prior can be interpreted as "virtual observations," hypothetical data points used to estimate specific properties~\cite{diaconis1979conjugate}. 

\begin{remark}  
In the NIG distribution, $\mu$ is estimated from $\nu$ virtual samples with a mean of $\gamma$, while $\sigma^2$ is inferred from $\alpha$ virtual samples with a sum of squared deviations proportional to $\nu$. Larger $\nu$ and $\alpha$ indicate stronger evidential support, reducing uncertainty in the prior.  
\end{remark}

This interpretation allows us to quantify the confidence in the reconstructed feature distribution by linking the hyperparameters $\gamma$, $\nu$, $\alpha$, and $\beta$ to virtual observations. Using the NIG distribution, we define two types of uncertainty:
\begin{itemize}[leftmargin=*]
    \item \textbf{Reconstruction Uncertainty:} This measures the variance of $\mu$ in $\mathcal{N}_0$, reflecting the model's uncertainty in reconstruction, defined as $\mathcal{U}_{\text{reconst}}^f = \frac{\beta}{\nu (\alpha - 1)}$.
    \item \textbf{Graph Uncertainty:} This represents the expected value of $\sigma^2$ in $\mathcal{N}_0$, expressed as $\mathcal{U}_{\text{graph}}^f = \frac{\beta}{\alpha - 1}$, capturing inherent graph randomness.
\end{itemize}
These uncertainties reflect both the data's inherent randomness and the model’s ability to reliably reconstruct it.

\subsubsection{Topological Evidence}
For topology reconstruction, traditional decoders use a sigmoid function to predict the presence of an edge $e_{ij}$ between nodes $v_i$ and $v_j$, providing a point estimate in the range $[0, 1]$. However, this approach fails to quantify prediction uncertainty. To address this, we model the existence of edge $e_{ij}$ using a Bernoulli distribution $\mathcal{B}(p_{ij})$, where $p_{ij}$ is the edge's probability, and $\boldsymbol{p}$ represents the probability of all edges in $\mathcal{E}$ existing.

We adopt Subjective Logic (SL)~\cite{DBLP:books/sp/Josang16} to quantify uncertainty in the Bernoulli distribution. SL extends Dempster-Shafer Theory (DST)~\cite{DBLP:series/sfsc/Dempster08b} by using a Dirichlet distribution to formalize belief assignments, enabling a rigorous application of evidential theory to quantify evidence and uncertainty.
Following CEDL~\cite{DBLP:conf/iccvw/0001RR023}, we leverage SL to define a probabilistic framework for topological reconstruction based on topological evidence.

\begin{proposition}
A Beta distribution parameterized by evidence variables can represent the density of probability assignments for discrete outcomes, effectively modeling both second-order probabilities and associated uncertainties.
\end{proposition}

To disentangle reconstruction results from uncertainty, we introduce two positive evidence variables, $\mathbf{E}_{ij}$ and $\bar{\mathbf{E}}_{ij}$, for each potential edge $(i, j)$, where $\mathbf{E}_{ij}$ represents the evidence supporting the edge's existence and $\bar{\mathbf{E}}_{ij}$ represents the evidence against it, with both being positive.
We propose to model the probabilities $p_{ij}$ as a Beta distribution:
{\small
\begin{equation}
\label{equ:Beta}
p(p_{ij} \mid \varepsilon_{ij}, \bar{\varepsilon}_{ij}) = \frac{\Gamma(\varepsilon_{ij} + \bar{\varepsilon}_{ij})}{\Gamma(\varepsilon_{ij}) \Gamma(\bar{\varepsilon}_{ij})} p_{ij}^{\varepsilon_{ij} - 1} (1 - p_{ij})^{\bar{\varepsilon}_{ij} - 1},
\end{equation}
}where $\varepsilon_{ij} = \mathbf{E}_{ij} + 1$ and $\bar{\varepsilon}_{ij} = \bar{\mathbf{E}}_{ij} + 1$ are the Beta parameters. 

The evidence strength for edge $(i, j)$, called Beta strength, is given by:
{\small
\begin{equation}
\label{equ:Beta_strength}
S_{ij} = \varepsilon_{ij} + \bar{\varepsilon}_{ij},
\end{equation}
}indicating that a larger $S_{ij}$ reflects more information about the edge, whether supporting or opposing its presence.
The predicted edge probability is $\hat{p}_{ij} = \frac{\varepsilon_{ij}}{S_{ij}}$.

To quantify the uncertainty in topological reconstruction, we define two complementary measures:
\begin{itemize}[leftmargin=*]
    \item \textbf{Reconstruction Uncertainty:} This measures the uncertainty from the reconstruction process, reflecting the model's confidence in predicting edge presence or absence. It is inversely proportional to the Beta strength $\mathcal{U}_{\text{reconst}}^t = \frac{1}{S_{ij}}$.

    \item \textbf{Graph Uncertainty:} This captures inherent uncertainty in the graph structure, arising from conflicting or imbalanced evidence about edge existence. It is defined as:
    {\small
   \begin{equation}
   \label{equ:U_topo_graph}
   \mathcal{U}_{\text{graph}}^t = (b_{ij} + \bar{b}_{ij})\left(1 - \frac{\left|b_{ij} - \bar{b}_{ij}\right|}{b_{ij} + \bar{b}_{ij}}\right),
   \end{equation}
   } where $b_{ij} = \frac{\mathbf{E}_{ij}}{S_{ij}}$ and $\bar{b}_{ij} = \frac{\bar{\mathbf{E}}_{ij}}{S_{ij}}$ represent the belief masses supporting and opposing edge $e_{ij}$, respectively. This measure reflects the balance between the evidence for and against the edge’s existence. Higher graph uncertainty indicates balanced evidence, while lower uncertainty occurs with a clear belief in the edge's presence or absence.
\end{itemize}

By incorporating evidence variables and leveraging the Beta distribution, this framework allows us to disentangle predictions and uncertainties, making the reconstruction process more robust to graph variability and reconstruction ambiguity.

\subsection{From the Evidence to Reconstruction}
We now integrate feature and topology reconstruction into a unified evidential framework utilizing the evidence learned from the latent embeddings $\mathbf{Z}$. Given the embeddings obtained from the GAE encoder in Section~\ref{sec:GAE}, we model the joint distribution of the reconstructed graph as:
{\small
\begin{equation}
\label{equ:joint}
p(\boldsymbol{\mu}, \boldsymbol{\sigma}^2, \mathbf{p} \mid \mathbf{Z}) = \prod_{i=1}^N \prod_{h=1}^D p(\mu_{ih}, \sigma_{ih}^2 \mid \mathbf{Z}_i) \prod_{i=1}^N \prod_{j=1}^N p(p_{ij} \mid \mathbf{Z}_i, \mathbf{Z}_j),
\end{equation}
}where $\boldsymbol{\mu}$ and $\boldsymbol{\sigma}^2$ represent the mean and variance of node features, and $\mathbf{p}$ denotes the edge existence probabilities. $p(\mu_{ih}, \sigma_{ih}^2 \mid \mathbf{Z}_i)$ models the distribution for the $h$-th feature of node $v_i$, and $p(p_{ij} \mid \mathbf{Z}_i, \mathbf{Z}_j)$ models edge existence.

We model uncertainty in feature and topology reconstructions using the NIG and Beta distributions (Equations~\ref{equ:NIG} and \ref{equ:Beta}). To parameterize these, we use two neural networks $f_{\theta_1}$ and $f_{\theta_2}$ to estimate the distribution parameters directly from $\mathbf{Z}$:
\begin{equation*}
[\boldsymbol{\gamma}, \boldsymbol{\nu}, \boldsymbol{\alpha}, \boldsymbol{\beta}] = f_{\theta_1}(\mathbf{Z}), \quad [\boldsymbol{\varepsilon}, \boldsymbol{\bar{\varepsilon}}] = f_{\theta_2}(\mathbf{Z}).
\end{equation*}

Here, $f_{\theta_1}$ outputs the parameters $\boldsymbol{\gamma}$ (mean), $\boldsymbol{\nu}$ (degree of freedom), $\boldsymbol{\alpha}$, and $\boldsymbol{\beta}$ (scale parameters) for the NIG distribution, and $f_{\theta_2}$ outputs the evidence parameters $\boldsymbol{\varepsilon}$ and $\boldsymbol{\bar{\varepsilon}}$ for the Beta distribution. Please refer to Appendix~\ref{app:network} for detailed implementation.

% To ensure compatibility with GAE-based anomaly detection methods, both networks are lightweight MLPs. For $f_{\theta_1}$, the MLP estimates the NIG parameters for each node: $[\gamma_{ih}, \nu_{ih}, \alpha_{ih}, \beta_{ih}] = f_{\theta_1}(\mathbf{Z}_i)$, where $\gamma_{ih}$ corresponds to the predicted mean of the $h$-th feature for node $v_i$, and $\nu_{ih}$, $\alpha_{ih}$, $\beta_{ih}$ capture the uncertainty in this prediction.
% For $f_{\theta_2}$, the MLP estimates the evidence parameters for edge existence using the concatenation of node pair embeddings: $[\varepsilon_{ij}, \bar{\varepsilon}_{ij}] = f_{\theta_2}(\text{concat}[\mathbf{Z}_i, \mathbf{Z}_j])$, where $\varepsilon_{ij}$ and $\bar{\varepsilon}_{ij}$ represent the evidence supporting and opposing edge $e_{ij}$, respectively. 
% The ReLU activation ensures non-negativity, and an offset of $1$ aligns the parameters with the Beta distribution requirements.

With the learned evidential parameters, we express higher-order distributions encapsulating both reconstruction and associated uncertainties. For feature reconstruction with the NIG distribution, the mean of $\mu_{ih}$ estimates $\mathbf{X}_{ih}$:
{\small
\begin{equation}
\textbf{Feature Reconstruction:} \quad \hat{\mathbf{X}}_{ih} = \mathbb{E}[\mu_{ih}] = \gamma_{ih}. 
\end{equation}
}

For topology reconstruction with the Beta distribution, the mean of $\mathbf{p}_{ij}$ estimates $\mathbf{A}_{ij}$:
{\small
\begin{equation}
\textbf{Topology Reconstruction:} \quad \hat{\mathbf{A}}_{ij} = \mathbb{E}[p_{ij}] = \frac{\varepsilon_{ij}}{\varepsilon_{ij} + \bar{\varepsilon}_{ij}},
\end{equation}
}

\subsection{Optimization}
We define the joint evidential distribution $p(\boldsymbol{\mu}, \boldsymbol{\sigma}^2, \mathbf{p} \mid \mathbf{Z})$ in Equ~\ref{equ:joint}. 
The optimization process is framed as a multi-task learning problem with two objectives: (1) maximizing model evidence to enhance reconstruction accuracy, and (2) minimizing evidence for incorrect predictions to enforce uncertainty when the model is wrong.

\subsubsection{Data Perturbation}
In unsupervised graph reconstruction, using a fixed graph for evidential learning can lead to overfitting, particularly in uncertainty modeling, undermining anomaly detection. To improve generalization and ensure reliable uncertainty estimates, we introduce perturbations to node features and graph topology during each training iteration. Specifically, we add Gaussian noise to node features: $\tilde{\mathbf{x}}_i = \mathbf{x}_i + \mathbf{n}_i$, where $\mathbf{n}_i \sim \mathcal{N}(0, \sigma^2 \mathbf{I})$. For topology, we apply dropout to the adjacency matrix using a random mask $\mathbf{M}$, where $M_{ij} \sim \text{Bernoulli}(1 - p)$, resulting in the perturbed adjacency matrix $\tilde{\mathbf{A}} = \mathbf{A} \odot \mathbf{M}$. These perturbations expose the model to diverse data variations, mitigating overfitting and improving uncertainty estimation for anomaly detection.
% In unsupervised graph reconstruction, using a fixed graph for evidential learning can lead to overfitting, especially in uncertainty modeling, resulting in unreliable anomaly detection. To improve generalization and ensure trustworthy uncertainty estimates, we introduce perturbations to node features and graph topology during each training iteration. Specifically, we add Gaussian noise to the node features: $\tilde{\mathbf{x}}_i = \mathbf{x}_i + \mathbf{n}_i$, where $\mathbf{n}_i \sim \mathcal{N}(0, \sigma^2 \mathbf{I})$. For topology, we apply dropout to the adjacency matrix using a random mask $\mathbf{M}$, where $M_{ij} \sim \text{Bernoulli}(1 - p)$, and compute the perturbed adjacency matrix as $\tilde{\mathbf{A}} = \mathbf{A} \odot \mathbf{M}$. These perturbations expose the model to diverse data variations, preventing overfitting and enhancing uncertainty estimation for anomaly detection.

\subsubsection{Maximizing Reconstruction Fit}
To maximize the likelihood of the observed graph data, we utilize the evidential distributions for feature and topological reconstruction. For feature reconstruction, the negative log-likelihood (NLL) loss is defined as:
{\small
\begin{equation}
\label{equ:f_nll}
\begin{aligned}
\mathcal{L}^{\text{NLL}}_{f} &= \sum_{i=1}^N\sum_{h=1}^d \left( \frac{1}{2} \log\left(\frac{\pi}{\nu_{ih}}\right) - \alpha_{ih} \log(\Omega_{ih}) \right. \\
&\quad + \left. \left(\alpha_{ih} + \frac{1}{2}\right) \log\left((X_{ih} - \gamma_{ih})^2 \nu_{ih} + \Omega_{ih}\right) \right. \\
&\quad + \left. \log\left(\frac{\Gamma(\alpha_{ih})}{\Gamma\left(\alpha_{ih} + \frac{1}{2}\right)}\right) \right)
\end{aligned}
\end{equation}
}where $\Omega_{ih} = 2\beta_{ih}(1+\nu_{ih})$. Please refer to Appendix~\ref{app:NLL} for detailed derivation. 
Similarly, for topological reconstruction, the NLL loss is:
{\small
\begin{equation}
\mathcal{L}^{\text{NLL}}_{t} = \sum_{i=1}^N\sum_{j=1}^N \left( \mathbf{A_{ij}}\log(\frac{S_{ij}}{\varepsilon_{ij}})+(1-\mathbf{A_{ij}})\log(\frac{S_{ij}}{\bar{\varepsilon}_{ij}}) \right),
\end{equation}
}where $S_{ij}$ is the Beta strength as defined in Equ~\ref{equ:Beta_strength}.

\subsubsection{Minimizing Evidence for Errors}
To penalize incorrect reconstructions, we minimize evidence strength in regions with high error, preventing evidence from becoming excessively large, which could lead to NaN or Inf values during training. For topological reconstruction, this is achieved by minimizing the Kullback-Leibler (KL) divergence between the predicted Beta distribution and the non-informative prior $\text{Beta}(1, 1)$:
\begin{equation}
\mathcal{L}^{\text{R}}_{t} = \sum_{i=1}^N\sum_{j=1}^N  |\mathbf{A}_{ij} - \frac{\varepsilon_{ij}}{\varepsilon_{ij} + \bar{\varepsilon}_{ij}}| \cdot   \text{KL}[\text{Beta}(\varepsilon_{ij}, \bar{\varepsilon}_{ij}) \parallel \text{Beta}(1, 1)].
\end{equation}
For feature reconstruction, instead of using KL divergence with a zero-evidence prior, we penalize the predicted evidence based on reconstruction error:
\begin{equation}
\mathcal{L}^{\text{R}}_{f} = \sum_{i=1}^N \sum_{h=1}^d |X_{ih} - \gamma_{ih}| \cdot (2\nu_{ih} + \alpha_{ih}).
\end{equation}
This loss discourages large evidence values when prediction errors are high, promoting increased uncertainty in regions with poor predictions.

\subsubsection{Multi-task Learning Framework}
We combine the two objectives into a unified multi-task learning framework, with the total loss defined as::
\begin{equation}
\mathcal{L} = \lambda_1 \mathcal{L}^{\text{NLL}}_{f} + \lambda_2 \mathcal{L}^{\text{NLL}}_{t} + \lambda_3 \mathcal{L}^{\text{R}}_{f} + \lambda_4 \mathcal{L}^{\text{R}}_{t},
\end{equation}
where $\lambda_1, \lambda_2, \lambda_3, \lambda_4$ are hyperparameters balancing the contribution of each objective. This framework optimizes both feature and topology reconstruction while accounting for uncertainty in both.

\subsection{Anomaly Detection}
\label{sec:Anomaly Detection}
We define the anomaly score of a node $v$ based on uncertainties from the GAE, where a higher score indicates the node is more likely to be anomalous. To improve robustness, we combine uncertainty with reconstruction error. Given modality heterogeneity, GEL introduces parameters $\lambda_{\text{g}}$, $\lambda_{\text{r}}$, $\lambda_{\text{f}}$, and $\lambda_{\text{t}}$ to balance these factors. The anomaly score $y_v$ is:
\begin{equation*}
\label{equ:AS}
\begin{aligned}
y_v = & \lambda_{\text{f}}(\lambda_{\text{g}}\mathcal{U}_{\text{graph}}^f+\lambda_{\text{r}}\mathcal{U}_{\text{reconst}}^f) +\lambda_{\text{t}}(\lambda_{\text{g}}\mathcal{U}_{\text{graph}}^t + \lambda_{\text{r}}\mathcal{U}_{\text{reconst}}^t)\\ 
& + |\mathbf{X}_v - \hat{\mathbf{X}}_v| + \sum_{j \in N(v)}|\mathbf{A}_{vj} - \hat{\mathbf{A}}_{vj}|, 
\end{aligned}
\end{equation*}
where $N(v)$ denotes the neighbors of $v$.

\begin{figure}[t]
    \centering
    \begin{subfigure}[b]{0.23\textwidth}
        \centering
        \includegraphics[width=\textwidth]{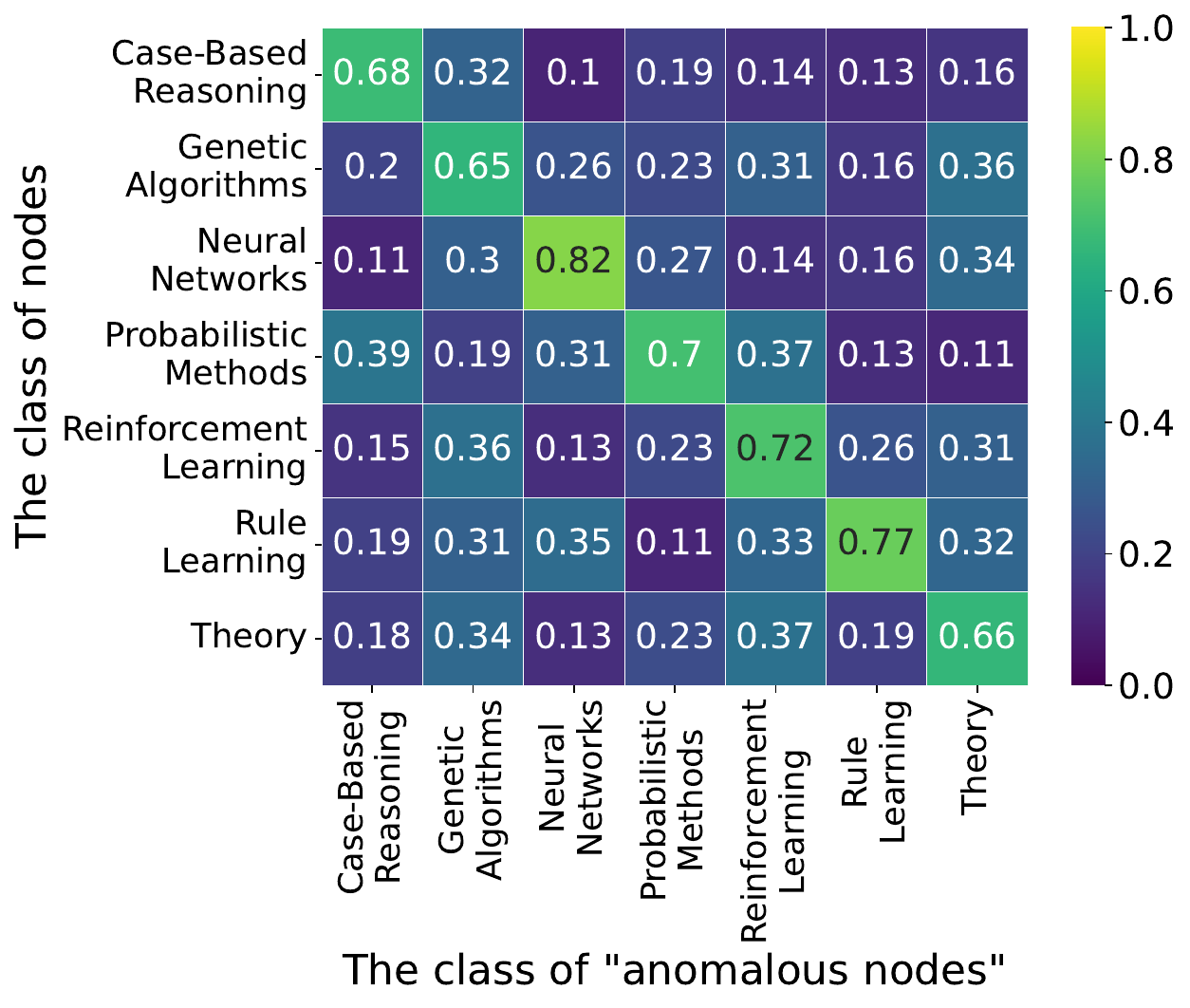}
        \caption{GEL: uncertainty estimates. }
        \label{fig:heatmap_a}
    \end{subfigure}
    \hfill % 添加一些水平间距
    \begin{subfigure}[b]{0.23\textwidth}
        \centering
        \includegraphics[width=\textwidth]{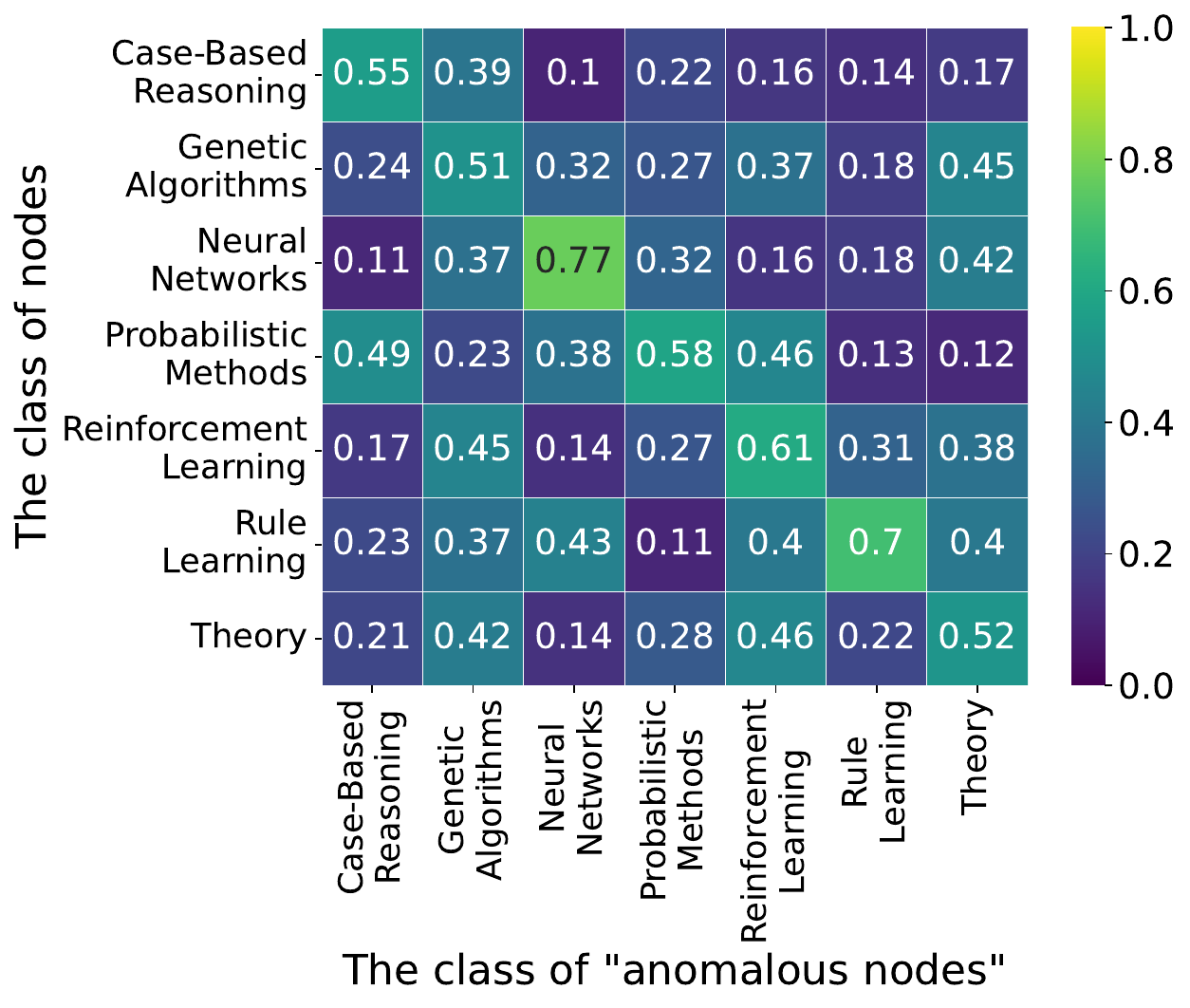}
        \caption{GCNAE: reconstruction error. }
        \label{fig:heatmap_b}
    \end{subfigure}
    \vspace{-1ex}
    \caption{Heatmaps of average normalized anomaly scores on the Cora dataset. Rows correspond to the class omitted during training (anomalous class), columns represent the class labels during testing.}
    \vspace{-1ex}
    \label{fig:heatmap}
\end{figure}

\subsection{Analysis}
To validate the effectiveness of uncertainty quantification in graph anomaly detection, we conduct experiments on the Cora dataset to test two hypotheses: (1) uncertainty measures are effective for identifying anomalies, and (2) incorporating uncertainty improves detection performance. We trained the Graph Evidential Learning (GEL) model, excluding nodes from each of the seven classes to create artificial anomalies, and computed normalized anomaly scores based on the model's uncertainty estimates. 
The model assigned significantly higher uncertainty to anomalous nodes compared to in-distribution nodes (Figure~\ref{fig:heatmap_a}), confirming the first hypothesis. To test the second hypothesis, we compared GEL against a baseline reconstruction error-based model, GCNAE~\cite{DBLP:journals/corr/KipfW16a}. The results (Figure~\ref{fig:heatmap_b}) show that GEL’s uncertainty-based detection outperforms the baseline, which solely relies on reconstruction error. This highlights that traditional models may overfit and lack a mechanism to quantify uncertainty, while GEL’s evidential framework improves anomaly detection by reducing overfitting.

\section{Experiments}
\label{sec:experiment}
\subsection{Experimental Setup}

\paragraph{Datasets}
% We conducted extensive experiments on five publicly available and popular real-world graph datasets to evaluate the quality of GEL in anomaly point detection tasks. These datasets include Weibo, Reddit, Disney, Books, and Enron, with the details of the datasets shown in Table~\ref{tab:dataset_stats}.
We conduct experiments on five publicly available and popular real-world graph datasets to evaluate the effectiveness and generalizability of GEL in anomaly point detection tasks. These datasets span a diverse range of domains, including social media (Weibo and Reddit), entertainment (Disney and Books), and corporate communication (Enron), demonstrating the broad applicability of our approach. Please refer to Appendix~\ref{app:dataset} for more details. 

\begin{figure*}[!htbp]
\vspace{-1ex}
\centering
\begin{minipage}[t]{0.70\textwidth} % 左侧表格部分
    \centering
    \begin{minipage}[t]{\textwidth} % 第一个表格
        \centering
        \begin{table}[H]
        \caption{Results on Weibo, Reddit, Disney, Books, Enron datasets w.r.t AUC.}
        \centering
        \vspace{-3mm}
        \scalebox{0.78}{
        \begin{tabular}{l|c|c|c|c|c}
        \toprule
        \textbf{Algorithm} & \textbf{Weibo} & \textbf{Reddit} & \textbf{Disney} & \textbf{Books} & \textbf{Enron} \\
        \midrule
        LOF & 56.5 $\pm$ 0.0 (56.5) & 57.2 $\pm$ 0.0 (57.2) & 47.9 $\pm$ 0.0 (47.9) & 36.5 $\pm$ 0.0 (36.5) & 46.4 $\pm$ 0.0 (46.4) \\ %ac
        IF & 53.5 $\pm$ 2.8(57.5) & 45.2 $\pm$ 1.7(47.5) & 57.6 $\pm$ 2.9(63.1) & 43.0 $\pm$ 1.8(47.5) & 40.1 $\pm$ 1.4(43.1) \\
        MLPAE & 82.1 $\pm$ 3.6(86.1) & 50.6 $\pm$ 0.0(50.6) & 49.2 $\pm$ 5.7(64.1) & 42.5 $\pm$ 5.6(52.6) & 73.1 $\pm$ 0.0(73.1) \\
        SCAN & 63.7 $\pm$ 5.6(70.8) & 49.9 $\pm$ 0.3(50.0) & 50.5 $\pm$ 4.0(56.1) & 49.8 $\pm$ 1.7(52.4) & 52.8 $\pm$ 3.4(58.1) \\
        Radar & \underline{\textbf{98.9 $\pm$ 0.1(99.0)}} & 54.9 $\pm$ 1.2(56.9) & 51.8 $\pm$ 0.0(51.8) & 52.8 $\pm$ 0.0(52.8) & 80.8 $\pm$ 0.0(80.8) \\
        ANOMALOUS & \underline{\textbf{98.9 $\pm$ 0.1(99.0)}} & 54.9 $\pm$ 5.6(60.4) & 51.8 $\pm$ 0.0(51.8) & 52.8 $\pm$ 0.0(52.8) & 80.8 $\pm$ 0.0(80.8) \\
        GCNAE & 90.8 $\pm$ 1.2(92.5) & 50.6 $\pm$ 0.0(50.6) & 42.2 $\pm$ 7.9(52.7) & 50.0 $\pm$ 4.5(57.9) & 66.6 $\pm$ 7.8(80.1) \\
        DOMINANT & 85.0 $\pm$ 14.6(92.5) & 56.0 $\pm$ 0.2(56.4) & 47.1 $\pm$ 4.5(54.9) & 50.1 $\pm$ 5.0(58.1) & 73.1 $\pm$ 8.9(85.0) \\
        DONE & 85.3 $\pm$ 4.1(88.7) & 53.9 $\pm$ 2.9(59.7) & 41.7 $\pm$ 6.2(50.6) & 43.2 $\pm$ 4.0(52.6) & 46.7 $\pm$ 6.1(67.1) \\
        AdONE & 84.6 $\pm$ 2.2(87.6) & 50.4 $\pm$ 4.5(58.1) & 48.8 $\pm$ 5.1(59.2) & 53.6 $\pm$ 2.0(56.1) & 44.5 $\pm$ 2.9(53.6) \\
        AnomalyDAE & 91.5 $\pm$ 1.2(92.8) & 55.7 $\pm$ 0.4(56.3) & 48.8 $\pm$ 2.2(55.4) & 62.2 $\pm$ 8.1(73.2) & 54.3 $\pm$ 11.2(69.1) \\
        GAAN & 92.5 $\pm$ 0.0(92.5) & 55.4 $\pm$ 0.4(56.0) & 48.0 $\pm$ 0.0(48.0) & 54.9 $\pm$ 5.0(61.9) & 73.1 $\pm$ 0.0(73.1) \\
        GUIDE & OOM\_C & OOM\_C & 38.8 $\pm$ 8.9(52.5) & 48.4 $\pm$ 4.6(63.5) & OOM\_C \\
        CONAD & 85.4 $\pm$ 14.3(92.7) & 56.1 $\pm$ 0.1(56.4) & 48.0 $\pm$ 3.5(53.1) & 52.2 $\pm$ 6.9(62.9) & 71.9 $\pm$ 4.9(84.9) \\

        G3AD & 95.1 $\pm$ 1.35(96.5) & \underline{62.1 $\pm$ 0.22(63.1)} & 65.3 $\pm$ 1.7(67.6) & 54.0 $\pm$ 4.2(58.7) & 72.39 $\pm$ 2.9(75.4) \\
        % BGNN & 81.4 $\pm$ 2.1(83.7) & 57.2 $\pm$ 1.9(60.3) & 58.9 $\pm$ 3.1(62.3) & 64.9 $\pm$ 1.8(66.9) & 69.2 $\pm$ 2.8(72.6) \\
        MuSE & 89.7 $\pm$ 4.1(95.3) & 53.7 $\pm$ 3.0(57.1) & 67.3 $\pm$ 1.5(69.1) & 64.3 $\pm$ 2.1(66.8) & 64.0 $\pm$ 3.77(68.2) \\
        OCGIN & 73.2 $\pm$ 2.8(76.2) & 51.8 $\pm$ 1.9(53.4) & 56.1 $\pm$ 1.5(57.7) & 64.4 $\pm$ 2.5(67.8) & 54.1 $\pm$ 1.8(56.5) \\
        GAD-NR & 87.71 $\pm$ 5.39(92.09) & 57.99 $\pm$ 1.67(59.90) & \underline{76.76 $\pm$ 2.75(80.03)} & \underline{65.71 $\pm$ 4.98(69.79)} & \underline{80.87 $\pm$ 2.95(82.92)} \\
        \midrule
        % \hline
        \textbf{GEL} & 89.32$\pm$3.36 (92.92) & \textbf{62.76$\pm$2.42(64.20)} & \textbf{78.21$\pm$4.94(82.48)} &\textbf{70.79$\pm$3.16(74.20)} & \textbf{82.34$\pm$2.93(85.76)} \\
        \bottomrule
        \end{tabular}}
        \label{tab:comparison_AUC}
        \end{table}
    \end{minipage}
    \begin{minipage}[t]{\textwidth}
        \centering
        \begin{table}[H]
        \vspace{-1em}
        \caption{Results on Weibo, Disney, Enron datasets w.r.t Recall@K.}
        \centering
        \vspace{-3mm}
        \scalebox{0.86}{
        \scriptsize 
        \setlength{\tabcolsep}{1pt}
        \begin{tabular}{l|ccccccccccc}
        \toprule
        \textbf{Dataset} & MLPAE & SCAN & Radar & GCNAE & DONE & AdONE & GAAN & GUIDE & CONAD & GAD-NR & \textbf{GEL} \\
        \midrule
        Weibo (Recall@500) & 51.82 $\pm$ 0.22 & 11.53 $\pm$ 0.00 & 53.60 $\pm$ 0.00 & 52.74 $\pm$ 0.60 & 42.94 $\pm$ 5.29 & 48.30 $\pm$ 3.09 & 53.14 $\pm$ 0.39 & OOM\_C & 26.28 $\pm$ 4.12 & \underline{60.22 $\pm$ 2.67} & \textbf{65.53 $\pm$ 5.23} \\
        Disney (Recall@50) & 32.86 $\pm$ 12.66 & 39.29 $\pm$ 0.00 & 39.29 $\pm$ 0.00 & 47.14 $\pm$ 12.04 & 43.57 $\pm$ 4.16 & 37.14 $\pm$ 5.80 & 40.71 $\pm$ 6.62 & 40.00 $\pm$ 2.67 & 21.43 $\pm$ 0.00 & \underline{67.85 $\pm$ 4.10} & \textbf{78.57 $\pm$ 2.70} \\
        Enron (Recall@1000) & 9.84 $\pm$ 4.11 & 7.38 $\pm$ 0.00 & 12.57 $\pm$ 0.00 & 8.20 $\pm$ 0.73 & 8.52 $\pm$ 4.19 & 2.13 $\pm$ 2.58 & 9.18 $\pm$ 1.31 & OOM\_C & 10.38 $\pm$ 0.00 & \underline{15.23 $\pm$ 1.41} & \textbf{20.00 $\pm$ 3.94} \\
        \bottomrule
        \end{tabular}}
        \label{tab:comparison_Recall@K}
        \end{table}
    \end{minipage}
\end{minipage} %
\hfill 
\begin{minipage}[t]{0.27\textwidth} % 右侧子图部分
    \centering
    \vspace{3em}
    \begin{figure}[H]
        \label{fig:overall}
        \begin{subfigure}[t]{1.0\textwidth}
            \centering
            \includegraphics[width=\linewidth]{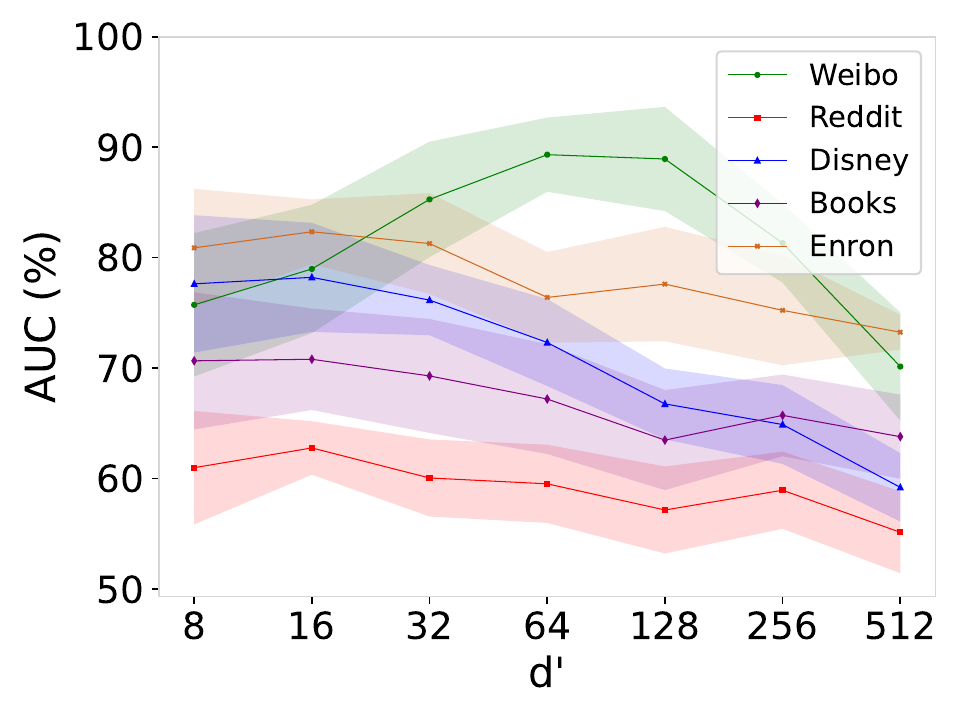}
        \end{subfigure}
        \vspace{-3em} % 减少一些垂直间距
        \begin{subfigure}[t]{1.0\textwidth}
            \centering
            \includegraphics[width=\linewidth]{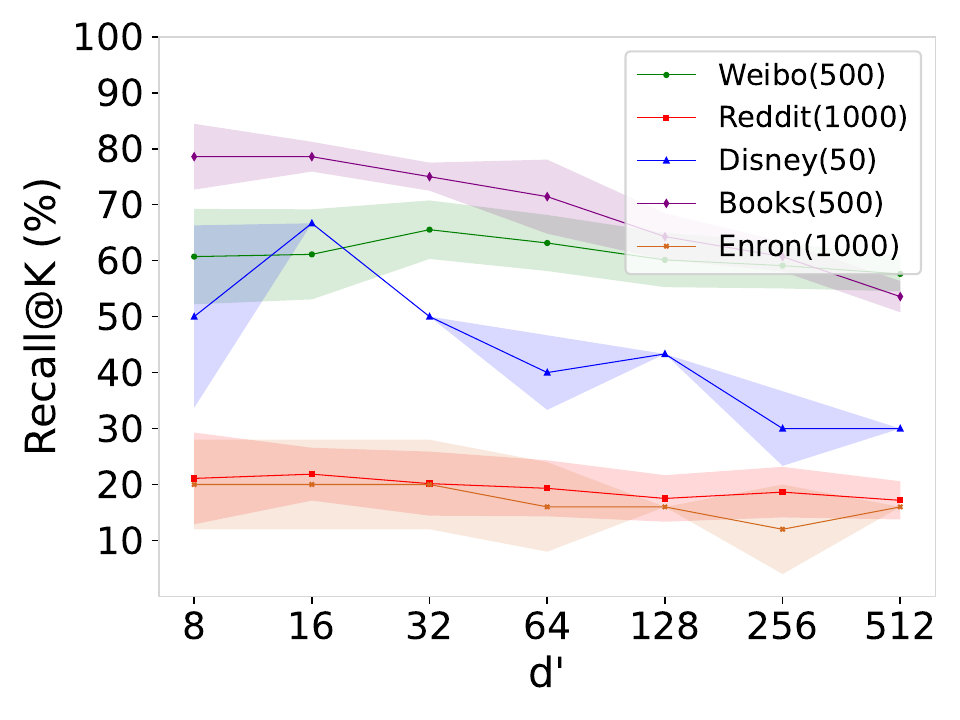}
        \end{subfigure}
        \vspace{1em}
        \caption{Impact of hidden layer dimension.}
    \label{fig:d_impact}
    \end{figure}
\end{minipage}
\end{figure*}

\paragraph{Baselines and Metrics}
Following BOND~\cite{DBLP:conf/nips/LiuDZDHZDCPSSLC22}, we select nineteen baseline anomaly detection models for comparison. The baseline models include non-graph-based methods such as LOF~\cite{DBLP:conf/sigmod/BreunigKNS00}, IF~\cite{DBLP:conf/aaai/BandyopadhyayLM19}, and MLPAE~\cite{DBLP:conf/pricai/SakuradaY14}, as well as clustering and matrix factorization-based anomaly detection algorithms like SCAN~\cite{DBLP:conf/kdd/XuYFS07}, Radar~\cite{DBLP:conf/ijcai/LiDHL17}, and ANOMALOUS~\cite{DBLP:conf/ijcai/PengLLLZ18}. Furthermore, we compared GEL with other models employing graph neural networks, such as AnomalyDAE~\cite{DBLP:conf/icassp/FanZL20}, GCNAE~\cite{kipf2016variational}, DOMINANT~\cite{ding2019deep}, GAD-NR~\cite{DBLP:conf/wsdm/RoySLYES024}, G3AD~\cite{bei2024guarding}, MuSE~\cite{kim2024rethinking}, OCGIN~\cite{zhao2023using}, and adversarial learning-based GAAN~\cite{DBLP:conf/cikm/ChenLWDLB20} and contrastive learning-based CONAD~\cite{DBLP:conf/pakdd/XuHZDL22}. For detailed description, please refer to Appendix~\ref{app:baseline}. 
Consistent with~\cite{DBLP:conf/wsdm/RoySLYES024, DBLP:journals/corr/abs-2409-09770}, we evaluate anomaly detection performance using the Area Under the ROC Curve (AUC) and Recall@K metrics~\footnote{The code is available on https://github.com/wuanjunruc/GEL.}.

\begin{figure}[!]
  \centering
  \includegraphics[width=0.8\linewidth]{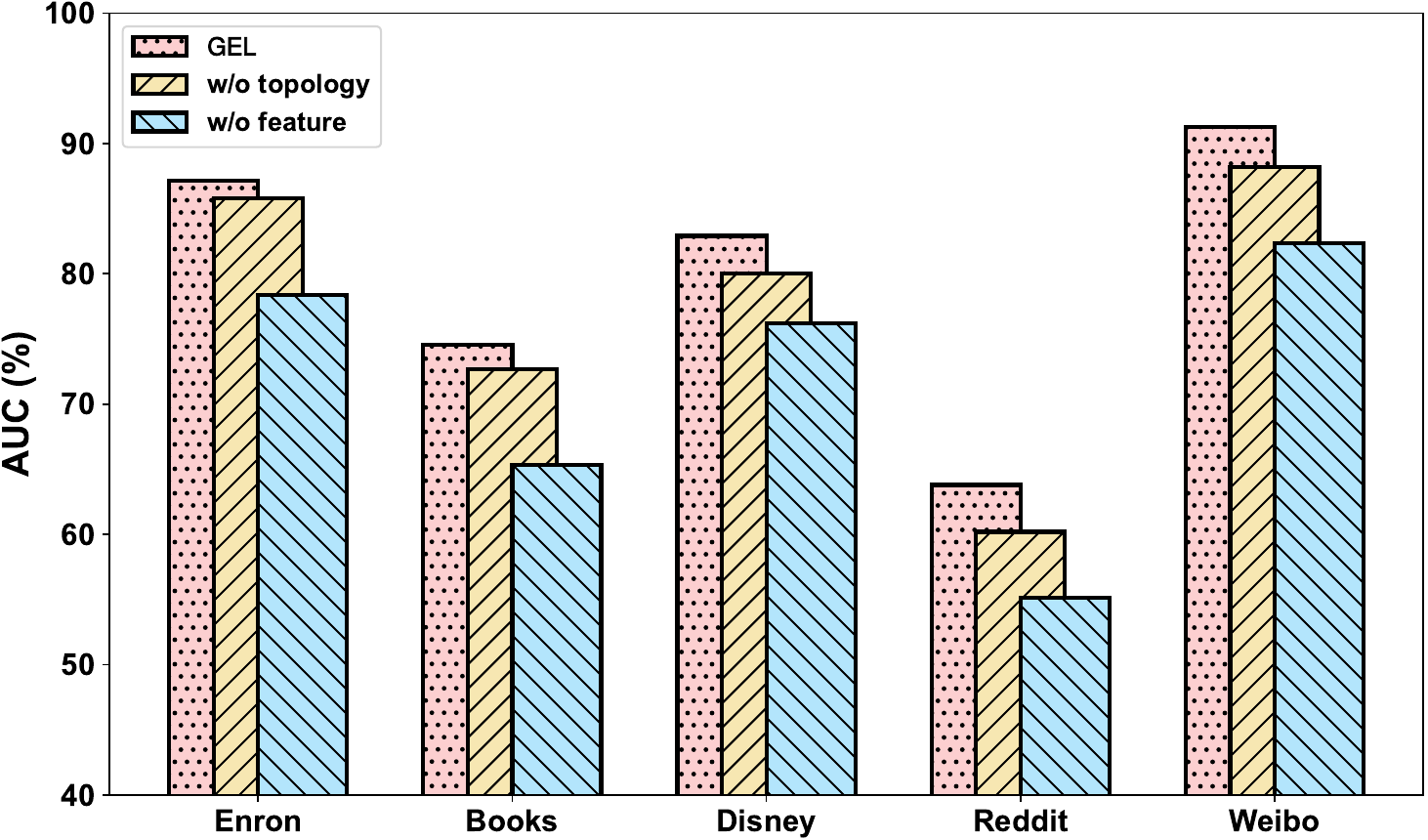}
  \vspace{-1ex}
  \caption{Impact of removing Different Modality.}
  \vspace{-1ex}
  \label{fig:ablation}
\end{figure}

\subsection{Performance Comparison (RQ1)}

We evaluate anomaly detection on five datasets. Best-performing models are highlighted in \textbf{bold}, and the best results among baselines are \underline{underlined}. `OOM\_C' indicates models that ran out of GPU memory. Detailed results are presented in Tables~\ref{tab:comparison_AUC} and~\ref{tab:comparison_Recall@K}. Key observations include:
\vspace{-1ex}
\begin{itemize}[leftmargin=*]
    \item GEL significantly outperforms other methods on four out of five datasets in both AUC and Recall@K. Specifically, GEL achieves an average AUC improvement of 4.64\% over the best baseline, demonstrating its effectiveness in capturing representational and structural information in graph data. By quantifying uncertainty in graph reconstruction, GEL uncovers hidden anomalous patterns and achieves robust anomaly detection.

    \item GEL shows substantial improvements on the Reddit and Books datasets, with AUC increases of 8.2\% and 7.7\%, respectively. The low anomaly rates (less than 3.5\%) in these datasets highlight GEL's precision in handling imbalanced data distributions.

    \item GAE-based methods (e.g., AnomalyDAE, DOMINANT, GCNAE, DONE) generally surpass traditional methods (LOF, IF, MLPAE, SCAN) due to their ability to learn key features and capture complex graph relationships. Traditional methods require manual feature selection, challenging in high-dimensional datasets, and often generalize poorly. For example, LOF detects anomalies based on local density but fails to identify attribute-based anomalies. Similarly, GEL employs graph reconstruction and quantifies uncertainty to effectively handle complex data structures and reveal inherent uncertainties.
\end{itemize}

\begin{figure*}[!t]
    % \centering
    \begin{subfigure}[b]{0.24\textwidth}
        \centering
        \includegraphics[width=\textwidth]{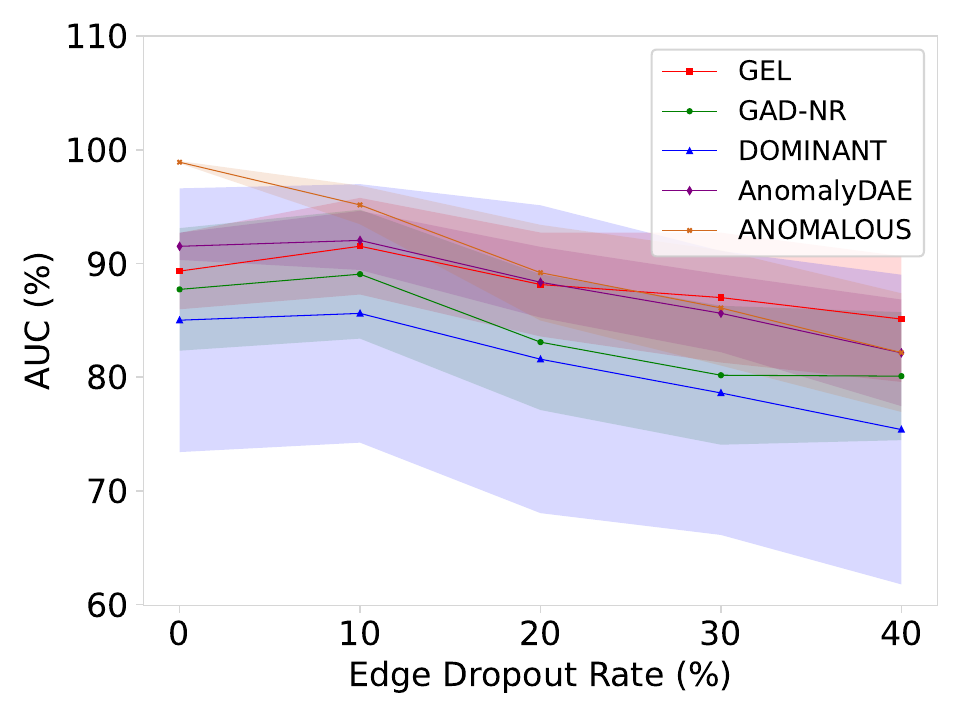}
        \caption{Weibo with dropout}
    \end{subfigure}
    \begin{subfigure}[b]{0.24\textwidth}
        \centering
        \includegraphics[width=\textwidth]{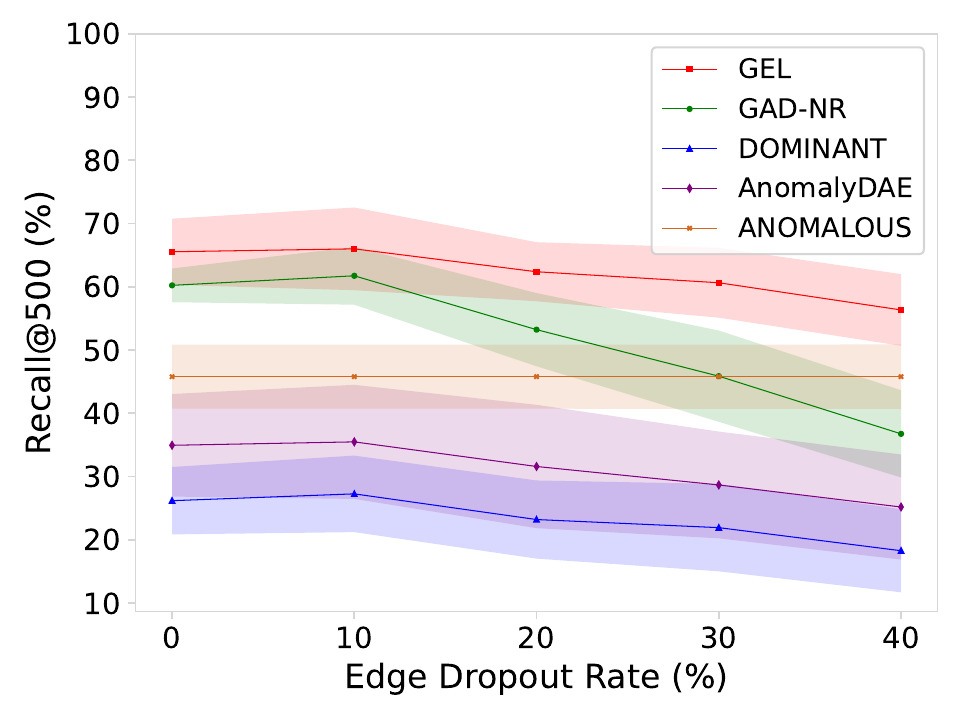}
        \caption{Weibo with dropout}
    \end{subfigure}
    \begin{subfigure}[b]{0.24\textwidth}
        \centering
        \includegraphics[width=\textwidth]{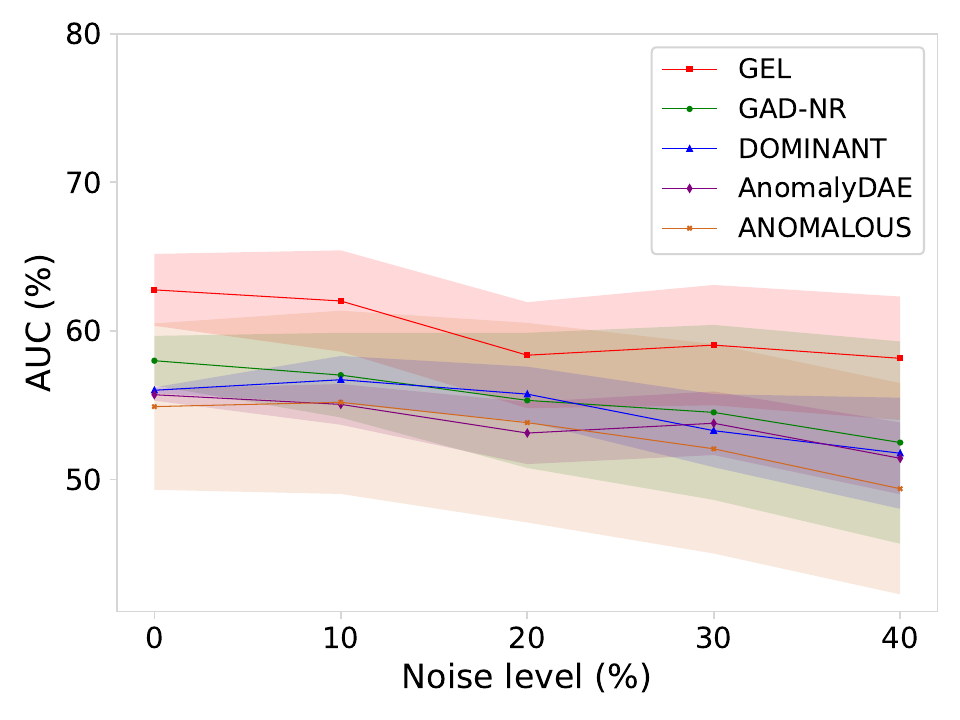}
        \caption{Reddit with noise}
    \end{subfigure}
    \begin{subfigure}[b]{0.24\textwidth}
        \includegraphics[width=\textwidth]{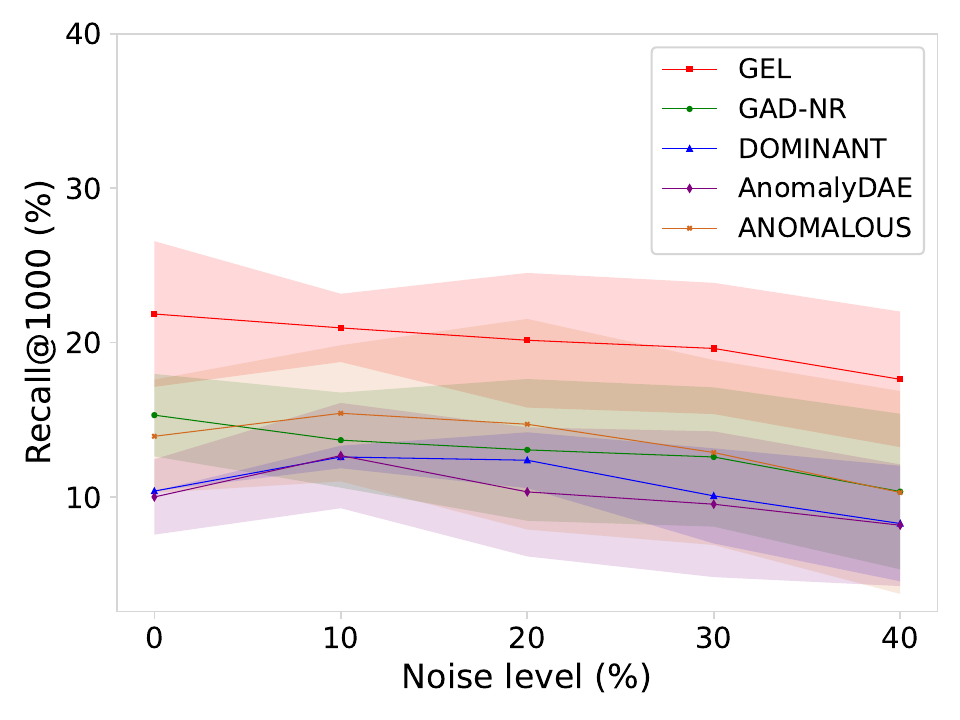}
        \caption{Reddit with noise}
    \end{subfigure}
    \vspace{-1ex}
    \caption{Changes of AUC and Recall@K as the level of data disturbance increases.}
    \vspace{-1ex}
    \label{fig:rob}
\end{figure*}

\subsection{Ablation Study}

Modality heterogeneity in graph reconstruction challenges GEL to integrate uncertainties from both node features and topological structures. To assess the impact of uncertainties in different modalities on GEL's performance, we conducted an ablation study with two variants:
1) \textbf{w/o feature}: GEL disregards uncertainty in node attributes by removing the feature evidence estimate, relying solely on topological structure uncertainty for anomaly detection.
2) \textbf{w/o topology}: GEL ignores uncertainty in the topological structure by removing the topology evidence estimate, using only node attribute uncertainty for anomaly detection. 

Figure~\ref{fig:ablation} presents our key observations:
1) GEL achieves optimal performance when uncertainties from both modalities are considered; removing any modality results in a noticeable performance decline.
2) The performance degradation is more pronounced when the feature modality is omitted, indicating that uncertainty in node attributes plays a critical role in anomaly detection. This suggests that node attributes provide richer information, enhancing the model's capability to detect anomalies.

% \subsection{Ablation Study (RQ2)}
% During graph reconstruction, modality heterogeneity introduces a challenge, prompting GEL to integrate the uncertainties associated with both node features and the topological structure. To assess the impact of uncertainty across different modalities on GEL's performance, we conducted an ablation study, introducing several variants: 1) \textbf{w/o feature:} In this configuration, GEL disregards the uncertainty related to node attributes—i.e., it removes the feature evidence estimate—relying solely on the uncertainty in the topological structure for anomaly detection. 2) \textbf{w/o topology:} Here, GEL solely incorporates the uncertainty in node attributes for topology anomaly evaluation, excluding the impact of uncertainty during the topological structure reconstruction, which removes the topology evidence estimate. 

% As shown in Figure~\ref{fig:ablation}, the following key observations were made: 1) GEL performs optimally when both modalities are fully utilized, and the removal of any modality leads to a noticeable decline in performance. 2) The performance degradation is especially pronounced when the feature modality is omitted, which may be due to the direct link between the uncertainty in node attributes and the identification of anomalies in the datasets used. Furthermore, node attributes potentially carry richer information, contributing to enhanced anomaly detection capabilities.

\subsection{Parameter study}
\begin{figure}[!t]
    % \centering
    \begin{subfigure}[b]{0.23\textwidth}
        \centering
        \includegraphics[width=\textwidth]{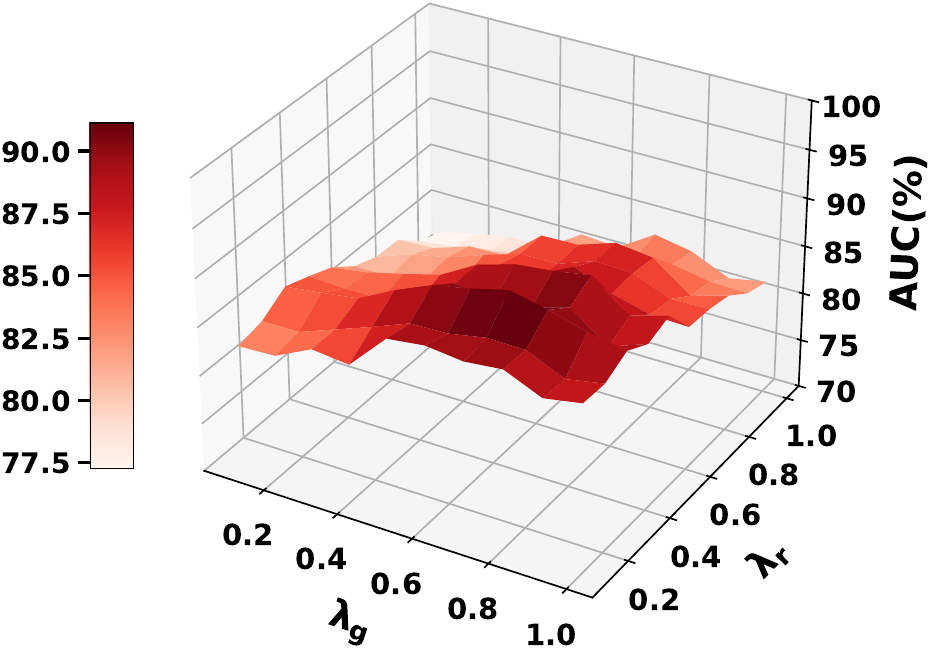}
        \caption{$\lambda_{r}$ and $\lambda_{g}$ on Weibo}
        \label{fig:lambda_rg_Weibo}
    \end{subfigure}
    \begin{subfigure}[b]{0.23\textwidth}
    \centering
    \includegraphics[width=\textwidth]{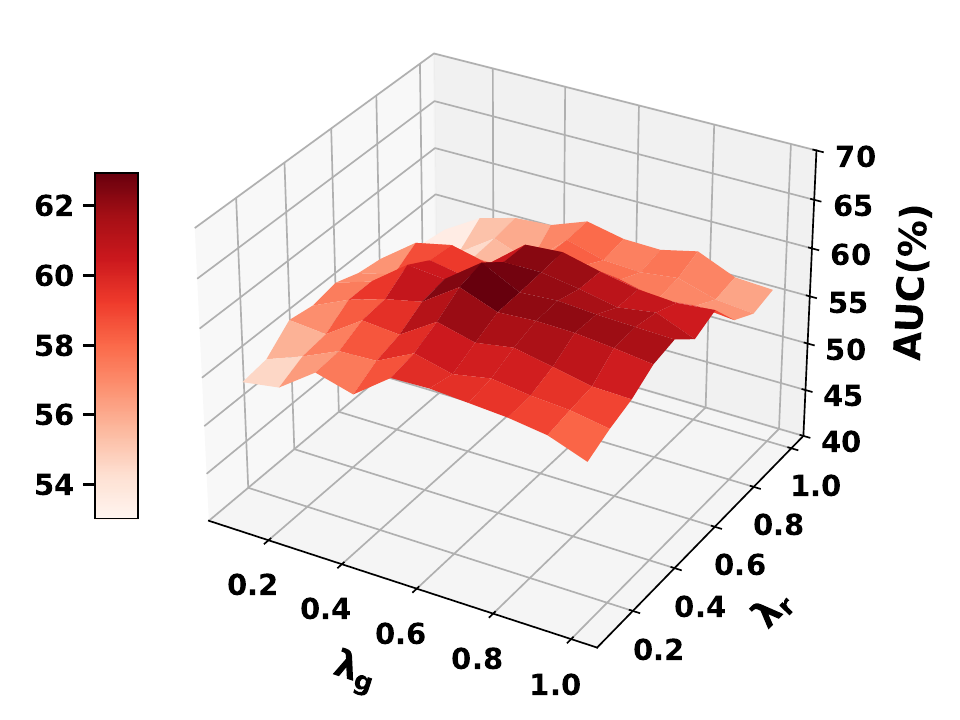}
    \caption{$\lambda_{r}$ and $\lambda_{g}$ on Reddit}
    \label{fig:lambda_rg_Reddit}
    \end{subfigure}
    
    \begin{subfigure}[b]{0.23\textwidth}
        \centering
        \includegraphics[width=\textwidth]{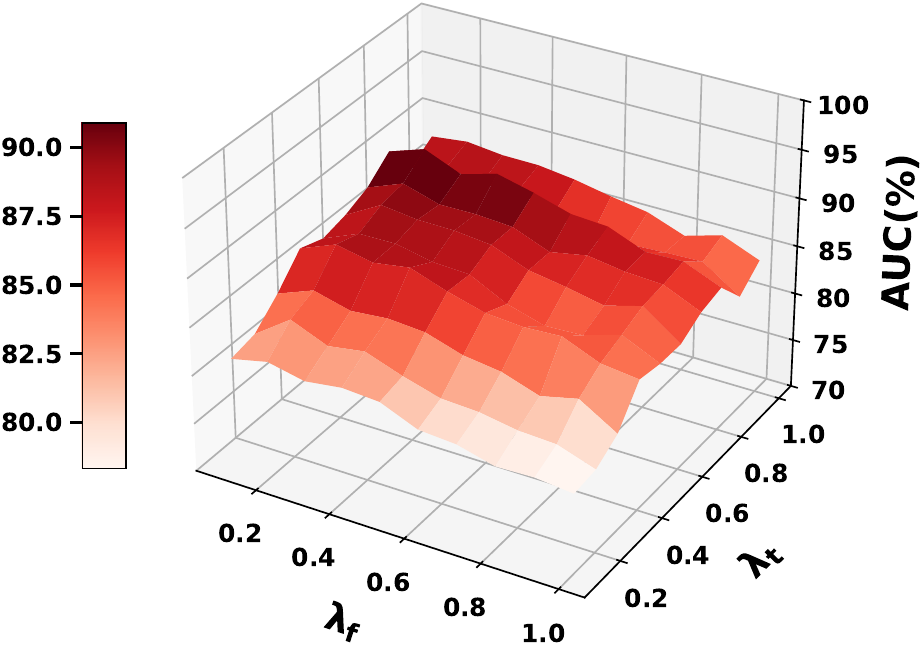}
        \caption{$\lambda_{f}$ and $\lambda_{t}$ on Weibo}
        \label{fig:lambda_ft_Weibo}
    \end{subfigure}
    \begin{subfigure}[b]{0.23\textwidth}
    % \centering
    \includegraphics[width=\textwidth]{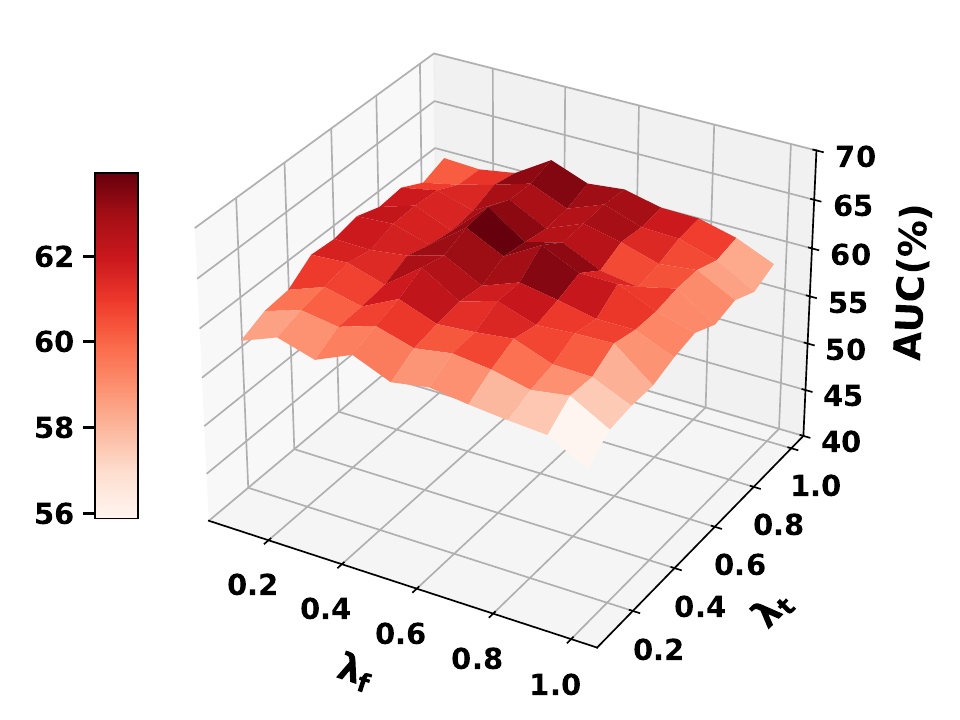}
    \caption{$\lambda_{f}$ and $\lambda_{t}$ on Reddit}
    \label{fig:lambda_ft_Reddit}
    \end{subfigure}
    \vspace{-1ex}
    \caption{Impacts of weights on diverse uncertainty.}
    \vspace{-3ex}
    \label{fig:4lambda}
\end{figure}
\subsubsection{Impact of Latent Dimension $d'$}

Figure~\ref{fig:d_impact} illustrates GEL's performance on the Weibo and Disney datasets as the latent dimension $d'$ increases. We observe that performance improves with larger $d'$, enabling the model to capture more reconstruction evidence and leading to a more accurate high-order evidential distribution, thus enhancing uncertainty quantification.

For the Weibo dataset, performance increases as $d'$ grows from 8 to 128; for the Disney dataset, it improves from 8 to 16. This difference likely stems from the varying node feature dimensions: Weibo has 400-dimensional features, whereas Disney has 28-dimensional features. Datasets with higher-dimensional node features require larger latent dimensions to capture richer evidence.

However, when $d'$ exceeds optimal values, performance declines due to overfitting. Overfitting is particularly detrimental in self-supervised models like GEL, as it leads to capturing noise and irrelevant details, reducing performance. GEL, based on a GAE architecture, aims to capture essential node information for uncertainty quantification; an excessively large $d'$ may hinder its generalization ability.

\subsubsection{Impact of $\lambda_{\text{r}}$ and $\lambda_{\text{g}}$}

As detailed in Section~\ref{sec:Anomaly Detection}, GEL models uncertainty through reconstruction uncertainty ($U_{\text{reconst}}$), reflecting the amount of evidence in reconstruction, and graph uncertainty ($U_{\text{graph}}$), representing the imbalance or adversarial nature of the evidence. We examine the impact of the weights $\lambda_{\text{r}}$ and $\lambda_{\text{g}}$ by adjusting them accordingly.

Figure~\ref{fig:4lambda}(a) and (b) illustrate that on the Weibo dataset, GEL achieves optimal performance when $\lambda_{\text{r}} = 0.7$ and $\lambda_{\text{g}} = 0.3$. This indicates that anomaly detection on Weibo relies more on reconstruction uncertainty, with moderate graph uncertainty aiding in handling adversarial evidence. A higher weight on reconstruction uncertainty allows the model to capture sufficient evidence for constructing a high-order evidential distribution, while a moderate weight on graph uncertainty helps manage the adversarial aspects effectively.
In contrast, on the Reddit dataset, the best performance occurs when $\lambda_{\text{r}} = 0.5$ and $\lambda_{\text{g}} = 0.5$, suggesting that equal consideration of both uncertainties is beneficial. This finding implies that the optimal weighting of uncertainties is influenced by dataset characteristics and anomaly patterns. Specifically, anomalies in Weibo involve insufficient evidence, emphasizing reconstruction uncertainty, whereas anomalies in Reddit are more affected by evidence conflicts.
These results demonstrate the flexibility and effectiveness of GEL in handling different anomaly patterns through the two types of uncertainty.

\subsubsection{Impact of $\lambda_{\text{f}}$ and $\lambda_{\text{t}}$}

We evaluated the effect of varying $\lambda_{\text{f}}$ and $\lambda_{\text{t}}$ on GEL's performance using the Weibo and Reddit datasets. As shown in Figure~\ref{fig:4lambda}(c) and (d), GEL achieves optimal performance on Weibo when $\lambda_{\text{f}} = 0.8$ and $\lambda_{\text{t}} = 0.2$, whereas on Reddit, the best performance occurs at $\lambda_{\text{f}} = 0.6$ and $\lambda_{\text{t}} = 0.4$.

These results indicate that a higher $\lambda_{\text{f}}$ makes anomaly detection more reliant on feature uncertainty estimates ($\boldsymbol{\gamma}$, $\boldsymbol{\nu}$, $\boldsymbol{\alpha}$, $\boldsymbol{\beta}$), which is beneficial for datasets with significant feature uncertainty differences. Conversely, a higher $\lambda_{\text{t}}$ emphasizes topology uncertainty estimates ($\boldsymbol{\varepsilon}$, $\boldsymbol{\bar{\epsilon}}$), aiding in scenarios with pronounced topology uncertainty differences.

The differing optimal weights suggest that the importance of feature and topology uncertainties varies across datasets, highlighting the need to balance $\lambda_{\text{f}}$ and $\lambda_{\text{t}}$ according to dataset characteristics.

\subsubsection{Impact of $\lambda_1 $, $\lambda_2$, $\lambda_3$ and $\lambda_4$}
We conducted experiments on multi-task training weights ($\lambda_1$, $\lambda_3$), setting $\lambda_2 = 1-\lambda_1$ and $\lambda_4 = 1-\lambda_3$ for efficiency. Results on Weibo show optimal performance at $\lambda_1 = 0.7$ and $\lambda_3 = 0.3$ (AUC: $89.22 \pm 3.21$), with similar patterns across other datasets.

\begin{table}[h]
\centering
\caption{AUC Results for Different $\lambda_1$ and $\lambda_3$ on Weibo}
\vspace{-4mm}
\scalebox{0.8}{
\begin{tabular}{c|ccccc}
\toprule
$\lambda_3 \backslash \lambda_1$ & 0.1 & 0.3 & 0.5 & 0.7 & 0.9 \\
\midrule
0.1 & $82.02 \pm 1.19$ & $84.51 \pm 3.17$ & $84.93 \pm 1.33$ & $86.14 \pm 0.22$ & $86.28 \pm 1.57$ \\
0.3 & $81.85 \pm 3.24$ & $83.93 \pm 2.59$ & $85.86 \pm 3.29$ & $\mathbf{89.22 \pm 3.21}$ & $88.49 \pm 2.54$ \\
0.5 & $79.76 \pm 2.72$ & $82.41 \pm 3.52$ & $85.02 \pm 2.43$ & $87.57 \pm 2.70$ & $88.24 \pm 0.83$ \\
0.7 & $80.09 \pm 2.08$ & $83.13 \pm 2.55$ & $82.69 \pm 1.95$ & $84.16 \pm 0.39$ & $85.11 \pm 3.01$ \\
0.9 & $79.22 \pm 1.21$ & $81.67 \pm 1.96$ & $83.57 \pm 2.85$ & $84.40 \pm 3.12$ & $83.61 \pm 2.47$ \\
\bottomrule
\end{tabular}}
\label{tab:lambda_results}
\vspace{-4mm}
\end{table}

\subsection{Robustness Study}

To evaluate robustness, we compared GEL with baselines under varying levels of noise and structural perturbations. Introducing increasing levels of noise to node features (Figure~\ref{fig:rob}(c) and (d)), we observed that while performance declined in all models due to their reliance on feature quality, GEL exhibited significantly less degradation than the baselines.

Similarly, we assessed the models under different proportions of edge dropout (Figure~\ref{fig:rob}(a) and (b)). At a dropout rate of 10\%, GEL and most baselines showed improved Recall@K, likely due to the regularization effect enhancing generalization~\cite{DBLP:conf/iclr/RongHXH20}. However, as dropout rates increased further, performance declined in all models. Notably, under higher dropout conditions, GEL's performance degradation was much smaller compared to GAD-NR.

GEL's robustness arises from its construction of high-order evidential and reconstruction distributions by capturing evidence from multiple dimensions to quantify uncertainty. This enables GEL to better handle missing or noisy data, maintaining relatively stable performance even under significant perturbations.

\subsection{Complexity Analysis}

GEL's computational complexity is comparable to standard GAE models with minimal additional overhead. We provide analysis from theoretical, implementation, and experimental perspectives.
From a theoretical perspective, GEL's complexity remains in line with standard GAEs:
\begin{itemize}
    \item Encoder: $O(E \cdot Z + N \cdot Z \cdot D)$ -- identical to standard GAEs.
    \item Feature evidence network ($f_{\theta_1}$): $O(N \cdot Z \cdot D)$ -- comparable to feature decoder in GAEs.
    \item Topology evidence network ($f_{\theta_2}$): $O(E \cdot Z \cdot D)$ -- comparable to topology decoder in GAEs.
    \item Uncertainty computation: $O(N \cdot D + E)$ -- negligible additional cost.
\end{itemize}
Where $N$ is the number of nodes, $E$ is the number of edges, $Z$ is the latent dimension, and $D$ is the feature dimension. The computational bottleneck remains in the encoder part, which is identical to standard GAE models.

The primary difference in GEL's decoder is outputting distribution parameters instead of point estimates, which adds negligible overhead in modern GPU-accelerated frameworks due to parallelization.

We compared training and inference times with the recent GAD-NR method based on GAE, which is a strong baseline. The results are shown in Tables~\ref{tab:train_time} and~\ref{tab:infer_time}.

\begin{table}[h]
\centering
\caption{Training Time Comparison}
\vspace{-3mm}
\scalebox{0.95}{
\begin{tabular}{l|ccccc}
\toprule
\textbf{Algorithm} & \textbf{Weibo} & \textbf{Reddit} & \textbf{Disney} & \textbf{Books} & \textbf{Enron} \\
\midrule
GAD-NR & 1m51s & 1m24s & 57s & 1m12s & 1m16s \\
GEL & 2m1s & 1m29s & 1m5s & 1m22s & 1m24s \\
\bottomrule
\end{tabular}
}
\label{tab:train_time}
\end{table}
\vspace{-5mm}

\begin{table}[h]
\centering
\caption{Inference Time Comparison}
\vspace{-3mm}
\scalebox{0.95}{
\begin{tabular}{l|ccccc}
\toprule
\textbf{Algorithm} & \textbf{Weibo} & \textbf{Reddit} & \textbf{Disney} & \textbf{Books} & \textbf{Enron} \\
\midrule
GAD-NR & 12s & 11.1s & 4.2s & 6.2s & 10.6s \\
GEL & 16.5s & 14.8s & 5.8s & 8.7s & 12.3s \\
\bottomrule
\end{tabular}
}
\label{tab:infer_time}
\vspace{2mm}
\end{table}

We also evaluated GEL on the large-scale DGraphFin dataset (3.7M nodes, 4.3M edges) and compared it with the GAD-NR. The results are shown in Table~\ref{tab:performance_DGraphFin}.

% \begin{table}[h]
% \centering
% \caption{Performance on DGraphFin Dataset}
% \vspace{-3mm}
% \begin{tabular}{l|ccccc}
% \toprule
% \textbf{Method} & \textbf{AUC} & \textbf{Recall@10000} & \textbf{Training} & \textbf{Inference} & \textbf{Memory} \\
% \midrule
% GAD-NR & $69.93$ & $45.46$ & 35m39s & 7m57s & 21.9GB \\
% GEL & $74.57$ & $51.34$ & 41m34s & 8m9s & 23.5GB \\
% \bottomrule
% \end{tabular}
% \label{tab:performance_DGraphFin}
% \end{table}

\begin{table}[h]
\centering
\caption{Performance on DGraphFin Dataset}
\vspace{-3mm}
\setlength{\tabcolsep}{2pt}
\scalebox{0.95}{ % 缩放到 80% 大小
\begin{tabular}{l|ccccc}
\toprule
\textbf{Method} & \textbf{AUC} & \textbf{Recall@10000} & \textbf{Training} & \textbf{Inference} & \textbf{Memory} \\
\midrule
GAD-NR & $69.93$ & $45.46$ & 35m39s & 7m57s & 21.9GB \\
GEL & $74.57$ & $51.34$ & 41m34s & 8m9s & 23.5GB \\
\bottomrule
\end{tabular}
}
\label{tab:performance_DGraphFin}
\vspace{-4mm}
\end{table}

These results show that GEL maintains a 6.64\% higher AUC than GAD-NR at million-node scale, requires only 17\% more training time, and operates within the memory constraints of a single NVIDIA RTX 3090 GPU (24GB).

\section{Conclusions and Limitations}
\label{sec:conclusion}
We introduced \emph{Graph Evidential Learning} (GEL), a novel framework that shifts graph anomaly detection from relying on reconstruction error to modeling uncertainty. GEL addresses the challenges of uncertainty diversity and modality heterogeneity by employing higher-order evidential distributions for both node features and topological structures. 
GEL quantifies both graph uncertainty and reconstruction uncertainty, enhancing robustness against noise and overfitting. Extensive experiments on benchmark datasets demonstrate that GEL achieves state-of-the-art performance.

While GEL advances graph anomaly detection through uncertainty modeling, our framework is currently oriented towards static graphs with fixed structures and attributes; extending GEL to dynamic graphs where topology and features evolve over time presents an avenue for future work. Lastly, although we employ specific evidential distributions tailored to continuous and discrete modalities, these choices may not capture all types of uncertainties in diverse datasets. Exploring alternative or more flexible evidential distributions could further enhance GEL's ability to detect a wider variety of anomalies.

\begin{acks}
This research was supported by the National Key R\&D Program of China (No. 2023YFC3304701), NSFC (No.62132017 and No.U2436209), the Shandong Provincial Natural Science Foundation (No.ZQ2022JQ32), the Beijing Natural Science Foundation (L247027), the Fundamental Research Funds for the Central Universities, the Research Funds of Renmin University of China, and the Young Elite Scientists Sponsorship Program by CAST under contract No. 2022QNRC001. It was also supported by Big Data and Responsible Artificial Intelligence for National Governance, Renmin University of China.
\end{acks}

\bibliographystyle{ACM-Reference-Format}
\balance
\bibliography{sample-base}

\appendix
\vspace{-1mm}
\section{Derivation of $\mathcal{L}_{f}^{\text{NLL}}$}
\label{app:NLL}
\vspace{-1mm}
In this subsection, we derive the negative log-likelihood loss for feature  (ie. Eq.~\ref{equ:f_nll}) of a NIG distribution. For convenience, we omit the subscript $_{ih}$ of the hyperparameters $\gamma, \upsilon, \alpha, \beta$ and denote them as as $\mathbf{m}$. We have: 
\vspace{-1mm}
\begin{align*}
& \quad  p(\mathbf{X}_{ih} | \mathbf{m}) \\
& = \int_{\mathbf{\theta}} p(\mathbf{X}_{ih}|\mathbf{\theta}) p(\mathbf{\theta} | \mathbf{m}) \, d\mathbf{\theta} \\
& = \int_{\sigma^2=0}^{\infty} \int_{\mu=-\infty}^{\infty} p(\mathbf{X}_{ih}|\mu, \sigma^2) p(\mu, \sigma^2 | \mathbf{m}) \, d\mu d\sigma^2 \\
& = \int_{\sigma^2=0}^{\infty} \int_{\mu=-\infty}^{\infty} p(\mathbf{X}_{ih}|\mu, \sigma^2) p(\mu, \sigma^2 | \gamma, \upsilon, \alpha, \beta) \, d\mu d\sigma^2 \\
& = 
\int_{\sigma^2=0}^{\infty} \int_{\mu=-\infty}^{\infty}
\left[
             \sqrt{\frac{1}{2 \pi \sigma^2}} \exp \left\{-\frac{(\mathbf{X}_{ih}-\mu)^{2}}{2\sigma^2}\right\}
\right] \\
& \quad \quad
\left[
 \frac{\beta^{\alpha}\sqrt{\upsilon}}{\Gamma(\alpha)\sqrt{2 \pi \sigma^{2}}} \left(\frac{1}{\sigma^{2}}\right)^{\alpha+1} \exp \left\{-\frac{2 \beta+\upsilon(\gamma-\mu)^{2}}{2 \sigma^{2}}\right\}
 \right]
 \, d\mu d\sigma^2 
 \\
& = \int_{\sigma^2=0}^{\infty}
\frac{\beta^\alpha \sigma^{-3-2\alpha}}{\sqrt{2\pi} \sqrt{1+1/\upsilon} \Gamma(\alpha)}
\exp \left\{ - \frac{2\beta + \frac{\upsilon (\mathbf{X}_{ih} - \gamma)^2}{1+\upsilon}}{2\sigma^2}\right\}
d\sigma ^2 
\\
& = \int_{\sigma=0}^{\infty}
\frac{\beta^\alpha \sigma^{-3-2\alpha}}{\sqrt{2\pi} \sqrt{1+1/\upsilon} \Gamma(\alpha)}
\exp \left\{ - \frac{2\beta + \frac{\upsilon (\mathbf{X}_{ih} - \gamma)^2}{1+\upsilon}}{2\sigma^2}\right\}
2\sigma d\sigma 
\\
& = \frac{\Gamma(1/2 + \alpha)}{\Gamma(\alpha) }  \sqrt{\frac{\upsilon}{\pi}}
\left(2\beta (1+\upsilon)\right)^\alpha
\left( \upsilon (\mathbf{X}_{ih}-\gamma)^2 + 2\beta (1+\upsilon)\right)^{-(\frac{1}{2}+\alpha)}, 
\end{align*} 
which is equivalent to: 
\begin{equation*}
p(\mathbf{X}_{ih} | \mathbf{m}) = \text{St}\left(\mathbf{X}_{ih}; \gamma, \frac{\beta(1+ \upsilon)}{\upsilon\,\alpha} , 2\alpha\right).
\end{equation*}

$\text{St}\left(y; \mu_\text{St}, \sigma_\text{St}^2, \upsilon_{St}\right)$ is the Student-t distribution evaluated at $y$ with location parameter $\mu_\text{St}$, scale parameter $\sigma_\text{St}^2$, and $\upsilon_{\text{St}}$ degrees of freedom. 
Afterwards, we can compute the negative log likelihood loss, $\mathcal{L}_{ih}$, for reconstruction $\mathbf{X}_{ih}$ as: 
\begin{align*}
\mathcal{L}_{ih} &= -\log p(\mathbf{X}_{ih}|\mathbf{m}) \\
&= -\log \left( \text{St}\left(\mathbf{X}_{ih}; \gamma, \frac{\beta(1+ \upsilon)}{\upsilon\,\alpha} , 2\alpha\right) \right),  \\
% &= \tfrac{1}{2}\log\left(\tfrac{\pi}{\upsilon}\right) - \alpha\log(2\beta(1+\upsilon)) + \left(\alpha+\tfrac{1}{2}\right) \log((y-\gamma)^2\upsilon + 2\beta(1+\upsilon)) + \log\left( \tfrac{\Gamma(\alpha)}{\Gamma(\alpha+\frac{1}{2})}\right) \\
\end{align*}
which has the following derivation:
\begin{align*}
\mathcal{L}_{ih} = & \tfrac{1}{2}\log\left(\tfrac{\pi}{\upsilon}\right) \\
& - \alpha\log(\Omega) + \left(\alpha+\tfrac{1}{2}\right) \log((y-\gamma)^2\upsilon + \Omega) \\
& + \log\left( \tfrac{\Gamma(\alpha)}{\Gamma(\alpha+\frac{1}{2})}\right)
\end{align*}
where $\Omega = 2\beta(1+\upsilon)$. 

\section{Baseline Methods}
\label{app:baseline}

\begin{itemize}[leftmargin=*]
\item \textbf{LOF}~\cite{DBLP:conf/sigmod/BreunigKNS00}: Local Outlier Factor measures node isolation by comparing its density to k-nearest neighbors using node features.
\item \textbf{IF}~\cite{DBLP:conf/aaai/BandyopadhyayLM19}: Isolation Forest uses decision trees, scoring anomalies by their proximity to the root.
\item \textbf{MLPAE}~\cite{DBLP:conf/pricai/SakuradaY14}: Multi-Layer Perceptron Autoencoder reconstructs node features, using reconstruction loss as the anomaly score.
\item \textbf{SCAN}~\cite{DBLP:conf/kdd/XuYFS07}: Structural Clustering Algorithm for Networks identifies clusters and labels structurally distinct nodes as anomalies.
\item \textbf{Radar}~\cite{DBLP:conf/ijcai/LiDHL17}: Radar detects anomalies using attribute residuals and structural coherence, with reconstruction residuals as the score.
\item \textbf{GCNAE}~\cite{kipf2016variational}: Graph Convolutional Network Autoencoder reconstructs node features and graph structure, using reconstruction error for anomaly detection.
\item \textbf{DOMINANT}~\cite{ding2019deep}: DOMINANT uses a two-layer GCN to reconstruct features and structure, with combined reconstruction error as the anomaly score.
\item \textbf{DONE}~\cite{DBLP:conf/wsdm/BandyopadhyayNV20}: DONE optimizes node embeddings and anomaly scores using MLPs in a unified framework.
\item \textbf{AnomalyDAE}~\cite{DBLP:conf/icassp/FanZL20}: AnomalyDAE reconstructs structure and attributes using a structural autoencoder and attribute decoder.
\item \textbf{GAAN}~\cite{DBLP:conf/cikm/ChenLWDLB20}: Graph Anomaly Adversarial Network uses a GAN framework, combining detection confidence and reconstruction loss for anomaly scoring.
\item \textbf{GUIDE}~\cite{DBLP:conf/bigdataconf/YuanZYHC021}: GUIDE uses motif-based degree vectors and mirrors DONE for anomaly detection.
\item \textbf{CONAD}~\cite{DBLP:conf/pakdd/XuHZDL22}: CONAD applies graph augmentation and contrastive learning for anomaly detection.
\item \textbf{G3AD}~\cite{bei2024guarding}: G3AD introduces an adaptive caching module to guard the GNNs from solely reconstructing the observed data that contains anomalies.
\item \textbf{MuSE}~\cite{kim2024rethinking}: MuSE uses the multifaceted summaries of reconstruction errors as indicators for anomaly detection.
\item \textbf{OCGIN}~\cite{zhao2023using}: OCGIN is an end-to-end graph outlier detection model that addresses the "performance flip" phenomenon by leveraging factors such as density disparity and overlapping support.

\item \textbf{GAD-NR}~\cite{DBLP:conf/wsdm/RoySLYES024}: Graph Anomaly Detection with Neighborhood Reconstruction enhances GAE by reconstructing a node’s full neighborhood, using reconstruction loss to detect anomalies.
\end{itemize}
For baseline methods, we used publicly available implementations and followed the parameter settings recommended in their original papers. Our reproduced results closely match those reported in their respective publications.

For fair comparison across all methods, we performed grid search over hyperparameters including:
\begin{itemize}
    \item Learning rate $\in \{0.0001, 0.001, 0.01\}$.
    \item Weight decay $\in \{0, 0.0001, 0.001\}$.
    \item Model-specific parameters following their respective papers.
\end{itemize}

\section{Dataset Overview}
\label{app:dataset}  
The details of these datasets are shown in Table~\ref{tab:dataset_stats}.
\begin{itemize}[leftmargin=*]  
\item \textbf{Weibo}~\cite{DBLP:conf/cikm/0003DYJW020}: A directed user interaction network from Tencent-Weibo. Anomalous users are identified by temporal post patterns. Features include post location and bag-of-words content representation.
\item \textbf{Reddit}~\cite{DBLP:conf/kdd/KumarZL19}: A network of user-subreddit interactions on Reddit. Banned users are anomalies. Features are LIWC-based vectors from user and subreddit posts.
\item \textbf{Disney}~\cite{DBLP:conf/icde/MullerSMB13} and \textbf{Books}~\cite{DBLP:conf/icdm/SanchezMLKB13}: Co-purchase networks for movies (Disney) and books (Books). Anomalies are based on student votes (Disney) or Amazon failure tags (Books). Features include price, review count, and ratings.
\item \textbf{Enron}~\cite{DBLP:conf/icdm/SanchezMLKB13}: An email interaction network. Spam-sending email addresses are anomalies. Features include average email length, recipient count, and email time intervals.

\end{itemize}

\begin{table}[ht]
\caption{Dataset statistics.}
\centering
\vspace{-3mm}
\scalebox{0.9}{
\begin{tabular}{*{6}{c}}
\toprule
\textbf{Statistic} & \textbf{Weibo} & \textbf{Reddit} & \textbf{Disney} & \textbf{Books} & \textbf{Enron} \\ \midrule
\#Nodes     & 8405    & 10984   & 124   & 1418  & 13533  \\
\#Edges      & 407963  & 168016  & 335   & 3695  & 176987 \\
\#Features   & 400     & 64      & 28    & 21    & 18     \\
\#Degree     & 48.5    & 15.3    & 2.7   & 2.6   & 13.1   \\
\#Anomalies  & 868     & 366     & 6     & 28    & 5      \\
Anomaly Ratio & 10.3\%  & 3.3\%   & 4.8\% & 2.0\% & 0.04\% \\ \bottomrule
\end{tabular}}
\label{tab:dataset_stats}
\vspace{-1ex}
\end{table}

% \section{Details about Metrics}
% \label{app:metric}

% \textbf{AUC:} The Area Under the Receiver Operating Characteristic Curve (AUC) measures the model's ability to distinguish between normal and anomalous instances across all decision thresholds. An AUC of 0.5 indicates random guessing, while an AUC close to 1.0 reflects excellent discrimination.

% \textbf{Recall@K:} Recall@K quantifies the proportion of true anomalies among the top K instances ranked by the model. It is particularly useful when resource constraints limit the number of instances that can be reviewed. Recall@K is calculated as: $\text{Recall@K} = \frac{T_K}{N_A}$, where $T_K$ is the number of true anomalies in the top K predictions, and $N_A$ is the total number of true anomalies.

\section{Details of Netowrk Implemetation}
\label{app:network}
To ensure compatibility with various GAE-based anomaly detection methods, we model $f_{\theta_1}$ and $f_{\theta_2}$ using lightweight Multi-Layer Perceptrons (MLPs). These networks are designed to output parameters that align with the prior definitions of their respective distributions. For $f_{\theta_1}(\cdot)$, we use the following architecture:
\begin{align}
\boldsymbol{\gamma} &= \mathbf{W}_3 \cdot \text{Tanh}(\mathbf{W}_2^{\gamma} \cdot \text{Tanh}(\mathbf{W}_1^{\gamma} \mathbf{Z} + \mathbf{b}_1^{\gamma}) + \mathbf{b}_2^{\gamma}) + \mathbf{b}_3^{\gamma}, \\
\boldsymbol{\nu} &= \text{ReLU}(\mathbf{W}_2^{\nu} \cdot \text{ReLU}(\mathbf{W}_1^{\nu} \mathbf{Z} + \mathbf{b}_1^{\nu}) + \mathbf{b}_2^{\nu}), \\
\boldsymbol{\alpha} &= \text{ReLU}(\mathbf{W}_2^{\alpha} \cdot \text{ReLU}(\mathbf{W}_1^{\alpha} \mathbf{Z} + \mathbf{b}_1^{\alpha}) + \mathbf{b}_2^{\alpha}) + 1, \\
\boldsymbol{\beta} &= \text{ReLU}(\mathbf{W}_2^{\beta} \cdot \text{ReLU}(\mathbf{W}_1^{\beta} \mathbf{Z} + \mathbf{b}_1^{\beta}) + \mathbf{b}_2^{\beta}).
\end{align}

For $f_{\theta_2}(\cdot)$, the evidence parameters $\varepsilon_{ij}$ and $\bar{\varepsilon}_{ij}$ for the existence and non-existence of edge $e_{ij}$ are computed as:
\begin{equation}
[\varepsilon_{ij}, \bar{\varepsilon}_{ij}] = \text{ReLU}(\mathbf{W}_2^{\varepsilon} \cdot \text{ReLU}(\mathbf{W}_1^{\varepsilon} \cdot \text{concat}([\mathbf{Z}_i, \mathbf{Z}_j]) + \mathbf{b}_1^{\varepsilon}) + \mathbf{b}_2^{\varepsilon}) + 1,
\end{equation}
where $\text{concat}([\mathbf{Z}_i, \mathbf{Z}_j])$ represents the concatenation of the latent embeddings of nodes $v_i$ and $v_j$, and the ReLU activation ensures the non-negativity of the evidence. The final "+1" term ensures consistency with the definition of the Beta parameters $\varepsilon_{ij} = \mathbf{E}_{ij} + 1$ and $\bar{\varepsilon}_{ij} = \bar{\mathbf{E}}_{ij} + 1$.

\end{document}